\documentclass[10pt,twocolumn,letterpaper]{article}

\usepackage[pagenumbers]{cvpr} % arXiv-style with page numbers

\usepackage{multirow}
\usepackage{placeins}   % \FloatBarrier (aaai2027 had it; cvpr doesn't)

\newcommand{\method}{FocusMem\xspace}

\usepackage{array}
\usepackage{tabularx}
\usepackage[most]{tcolorbox}
\usepackage{fvextra}   % verbatim prompt boxes with line breaking
\usepackage{float}     % [H] pins appendix figures in reading order
\newcommand{\sg}{\operatorname{sg}}
\newcommand{\pool}{\operatorname{Pool}}

\newcommand{\slot}[1]{\texttt{\{\{#1\}\}}}
\definecolor{tocblue}{RGB}{33,93,160}
\newcommand{\pdfcontentsdest}[1]{\hypertarget{#1}{}}
\newcommand{\pdfcontentslink}[2]{\hyperlink{#1}{\textcolor{tocblue}{#2}}}
\newcommand{\pdfcontentsentry}[3]{%
  \noindent
  \pdfcontentslink{#1}{#2}\nobreak\hfill
  \pdfcontentslink{#1}{#3}\par
}
\newcommand{\pdfcontentssubentry}[3]{%
  \begingroup
  \leftskip=1.25em
  \parindent=0pt
  \pdfcontentslink{#1}{#2}\nobreak\hfill
  \pdfcontentslink{#1}{#3}\par
  \endgroup
}

\newcounter{focusalgorithm}
\newenvironment{focusalgorithm}[1]{%
  \refstepcounter{focusalgorithm}%
  \begin{figure}[H]
  \hrule height 0.8pt
  \vspace{2pt}
  \noindent\textbf{Algorithm \thefocusalgorithm: #1}\par
  \vspace{2pt}
  \hrule height 0.4pt
  \vspace{3pt}
  \small
  \renewcommand{\arraystretch}{1.08}
}{%
  \vspace{2pt}
  \hrule height 0.8pt
  \end{figure}
}

\newtcolorbox{wideprompt}[1]{
  colback=gray!4,
  colframe=black!55,
  boxrule=0.6pt,
  arc=1mm,
  title={#1},
  fonttitle=\bfseries
}

\newtcolorbox{promptbox}[1]{
  breakable,
  colback=gray!4,
  colframe=black!55,
  boxrule=0.6pt,
  arc=1mm,
  title={#1},
  fonttitle=\bfseries
}

\definecolor{cvprblue}{rgb}{0.21,0.49,0.74}
\usepackage[pagebackref,breaklinks,colorlinks,allcolors=cvprblue]{hyperref}

\def\paperID{0000} % *** Enter the Paper ID here ***
\def\confName{CVPR}
\def\confYear{2026}

\title{FocusMem: Factorizing Content, Readout, and Trust in Latent GUI Memory}

\author{\bfseries
    Zhuoran Zhang\textsuperscript{1,2},
    Bowen Li\textsuperscript{1,2},
    Jingcheng Ju\textsuperscript{3},
    Yang Shi\textsuperscript{1},
    Qixun Wang\textsuperscript{1},
    \\
    \bfseries
    Haotian Wang\textsuperscript{4},
    Wei Chen\textsuperscript{1,2},
    Tengjiao Wang\textsuperscript{1,2}\thanks{Corresponding Author}
    \\[3pt]
    \textsuperscript{1}School of Computer Science, Peking University
    \\
    \textsuperscript{2}Key Lab of High Confidence Software Technologies, Peking University
    \\
    \textsuperscript{3}Institute of Information Engineering, Chinese Academy of Sciences
    \\
    \textsuperscript{4}Department of Computer Science and Technology, Tsinghua University
    \\
    \href{https://github.com/stanley-ju/FocusMem}{\texttt{https://github.com/stanley-ju/FocusMem}}
}

\begin{document}
\maketitle

% !TeX root = ../AnonymousSubmission2027.tex
\begin{abstract}
GUI agents must remember both useful experience
from earlier tasks and unfinished progress in the current
interaction. Latent memory offers a compact solution by
compressing multimodal trajectories into a few continuous
tokens. Existing methods, however, usually map each trajectory
to one fixed memory block and train it mainly through next-action
supervision. This creates three practical problems: important
details may be lost during compression, the same memory block
must serve different decision stages, and irrelevant retrieved
trajectories may still mislead the agent. We introduce \textbf{\method}, which separates these responsibilities
within a compact latent-memory interface. A role-aware content
basis encourages episodic memory to retain reusable experience
and working memory to retain task progress. A state-conditioned
readout generates a decision-specific view of the same stored
evidence, while a lightweight trust gate can suppress memory
blocks that appear irrelevant to the current step. All components
are trained while the GUI policy remains frozen. Across five GUI-agent benchmarks, \method consistently
outperforms a fully matched action-only fixed-memory baseline
and prior latent memory adaptations. Further analysis shows that
semantic and functional supervision preserve complementary
information, state-conditioned readout is more robust as
surrounding trajectory context grows, and the trust gate reduces
the harm caused by injected irrelevant episodic evidence. These
results show that effective latent memory depends not only on
compressing past interaction, but also on what is retained, what
is exposed, and what is allowed.
\end{abstract}
\section{Introduction}

GUI agents~\citep{nguyen2025gui} require memory at two complementary temporal scales~\citep{hu2025memory,zhang2025survey}.
\emph{Episodic memory} supplies reusable evidence from completed tasks~\citep{shi2026androtmem}, while
\emph{working memory} maintains constraints and progress in the current episode~\citep{wu2025human}.
Replaying either source as full trajectories is costly: repeated screenshots and
action traces rapidly expand the policy context
~\citep{liu2023lost,he2024webvoyager,tian2025mmina}. Text memory is compact and
interpretable but can lose visual or procedural details that are difficult to
verbalize~\citep{zhao2024expel,wang2024awm,ouyang2025reasoningbank}. Structured multimodal
memory retains richer evidence, yet reinserting interleaved screenshots and text
still consumes many input tokens
~\citep{zeng2026mementogui,zhu2026hymem}. These trade-offs motivate latent
memory, which compresses multimodal trajectories into continuous tokens.

Trajectory-based latent memory has progressed from compact continuous
experience to a shared interface for episodic and working evidence
~\citep{wu2025comem,zhang2026memw}. Existing methods, however, typically map
each selected memory item to a state-independent latent block and optimize this
mapping mainly through the downstream action objective. Retrieval determines
which items enter the candidate set, but not what remains recoverable after
compression, which part of a selected item should be exposed under the current
decision, or whether the resulting block should affect the policy. These
questions are especially important because episodic and working memory serve
different roles, while a single trajectory may contain evidence useful at
multiple decision stages~\citep{li2025echotrail,fang2026memp}.

% !TeX root = ../AnonymousSubmission2027.tex
\begin{figure*}[!t]
\centering

\includegraphics[width=0.9\textwidth]{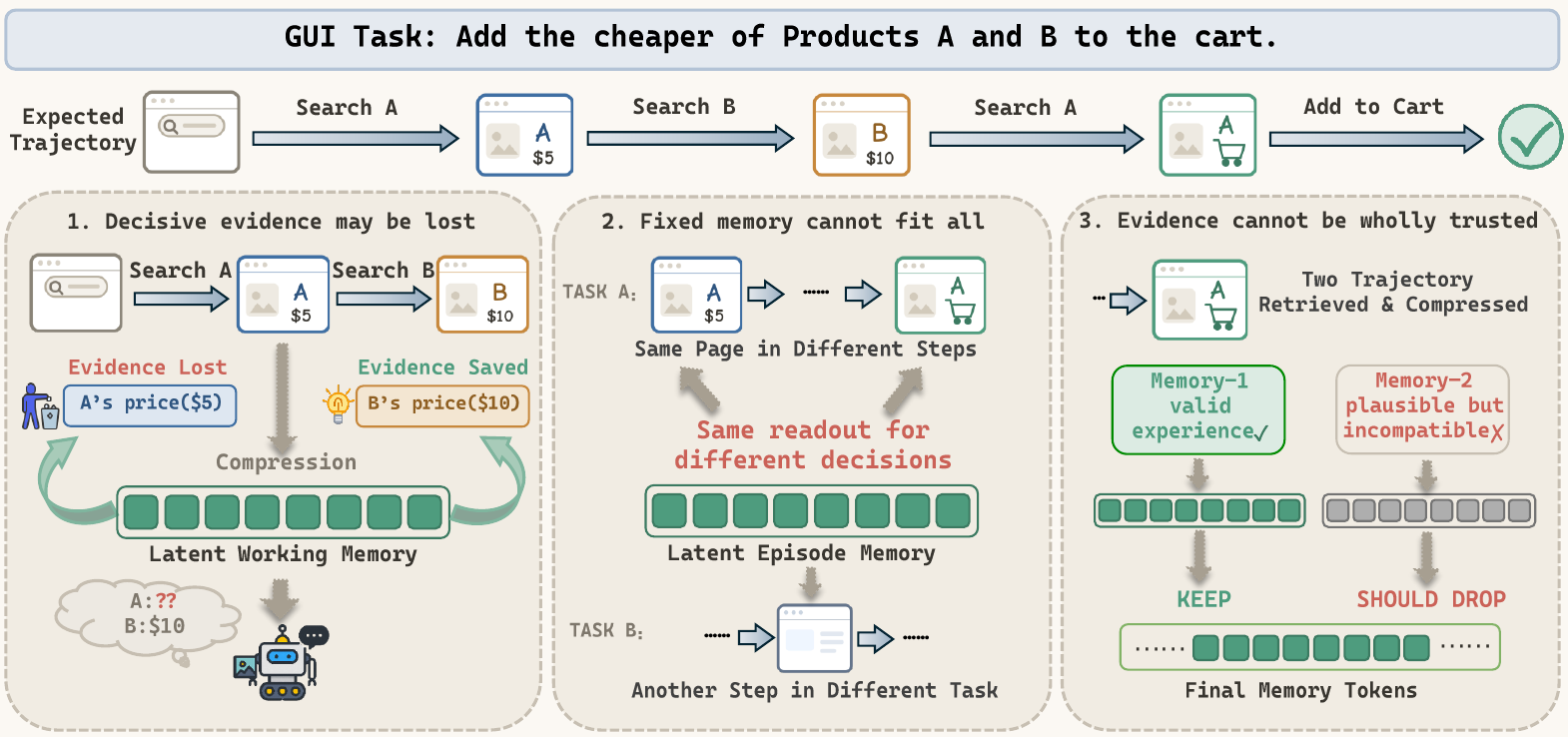}
\caption{Three latent-memory failure modes in one shopping trajectory. Compact
memory must preserve useful evidence, expose the decision-relevant part, and reject
misleading evidence.}
\label{fig:three-failures}
\end{figure*}
% !TeX root = ../AnonymousSubmission2027.tex

Figure~\ref{fig:three-failures} illustrates the resulting failure modes.
\emph{Content recoverability} concerns what must survive the bottleneck:
episodic memory should retain reusable experience, whereas working memory
should retain current constraints and progress. \emph{Conditional readout}
concerns how retained evidence is exposed: even when a decisive detail
survives compression, the same fixed block may serve interaction states that
require different parts of the trajectory. \emph{Evidence trust} concerns
whether selected evidence should enter the policy at all, since a retrieved
item may still be poorly matched to the current decision. 
Content,
readout, and trust are therefore distinct responsibilities rather than
interchangeable forms of memory selection.
Explicit memories can realize these distinctions by rewriting, selecting, or
dropping human-readable records. In latent memory, however, they must be learned
within the compact continuous interface itself.

We propose \textbf{\method}, which replaces the one-shot mapping from a memory
item to a fixed latent block with a factorized memory interface. Each episodic
or working item first learns a stable content basis suited to its role. The
current decision state then draws a compact, decision-specific view from the
same evidence, and a separate trust gate determines whether that view should
enter the policy. As a result, the same trajectory can provide different
information at different interaction stages, rather than relying on one fixed
summary for every decision, while mismatched evidence can be excluded under the
same bounded context budget.

We evaluate \textbf{\method} across five GUI-agent benchmarks under a common
experimental setup. It consistently improves task success over agents without
memory, raw-trajectory replay, and prior memory methods, while keeping the
policy-facing memory compact. To understand where these gains come from, we
conduct targeted analyses of the three responsibilities. We examine whether
useful information remains recoverable after compression, whether
state-conditioned readout can surface the decision-relevant part of the same
trajectory as more surrounding context is included under a fixed latent budget,
and whether irrelevant retrieved experience can be prevented from misleading
the agent. Together, the results show that the benefit comes from how latent
memory is formed, adapted to the current decision, and controlled before use,
rather than simply from providing more past interaction.

Our contributions are summarized as follows:
\begin{itemize}
    \item We formulate compact latent GUI memory around three distinct
    responsibilities: forming recoverable content, adapting its policy-facing
    view to the current decision, and controlling its influence on the policy.

    \item We introduce \textbf{\method}, a frozen-policy latent-memory framework
    that realizes these responsibilities through a role-aware content basis,
    state-conditioned readout, and an independent block-level trust gate.

    \item We provide matched evaluations across five GUI-agent benchmarks and
    targeted diagnostics that isolate the source of the gains and characterize
    when dynamic latent readout is most beneficial.
\end{itemize}

\FloatBarrier
% !TeX root = ../AnonymousSubmission2027.tex
\section{Related Work}

\paragraph{GUI-agent memory.}
GUI memory spans textual workflows and lessons
~\citep{cheng2025mga,wang2024awm,ouyang2025reasoningbank}, structured multimodal records
~\citep{zeng2026mementogui,zhu2026hymem,shi2025memoryvla}, history-aware reasoning
~\citep{wang2025historyawarereasoningguiagents}, and executable routine graphs
~\citep{qin2026executableagenticmemorygui,hu2025hiagent}.
Latent methods instead compress trajectories into policy-facing embeddings:
CoMEM~\citep{wu2025comem} encodes retrieved experience, while
Mem-W~\citep{zhang2026memw} uses a shared fixed compressor for episodic and
working memory. FocusMem retains this compact interface while separating
content formation, state-conditioned readout, and evidence trust.

\paragraph{Conditional and recoverable latent compression.}
InstructBLIP~\citep{dai2023instructblip} and
LaMem-VLA~\citep{qu2026lamemvla} use conditional queries for context-relevant
extraction, whereas Gist Tokens~\citep{mu2023gist},
ICAE~\citep{ge2023icae}, and
AutoCompressors~\citep{chevalier2023autocompressors} improve bottleneck
recoverability through reconstruction or distillation.
VisMem~\citep{yu2025vismem} and MemGen~\citep{zhang2025memgen} further study
adaptive latent memory beyond GUI trajectories. FocusMem combines recoverable
content supervision with state-conditioned generation for GUI
trajectories.

\paragraph{Learning and analyzing GUI memory.}
Recent work~\citep{choi2026naivevisualmemoryenough,liu2026memoryguiagentsreally,wang2026stamptrainingexplicitmemory}
studies what and when GUI agents should memorize, how passive records become
task-driving states, and failures of naive visual memory. These studies motivate
examining retained content, decision-time use, and harmful influence beyond
memory capacity. FocusMem studies these issues through content probes, evidence
expansion, and contamination analysis within a compact latent interface.

% !TeX root = ../AnonymousSubmission2027.tex
\begin{figure*}[!t]
\centering
\resizebox{0.98\textwidth}{!}{\includegraphics{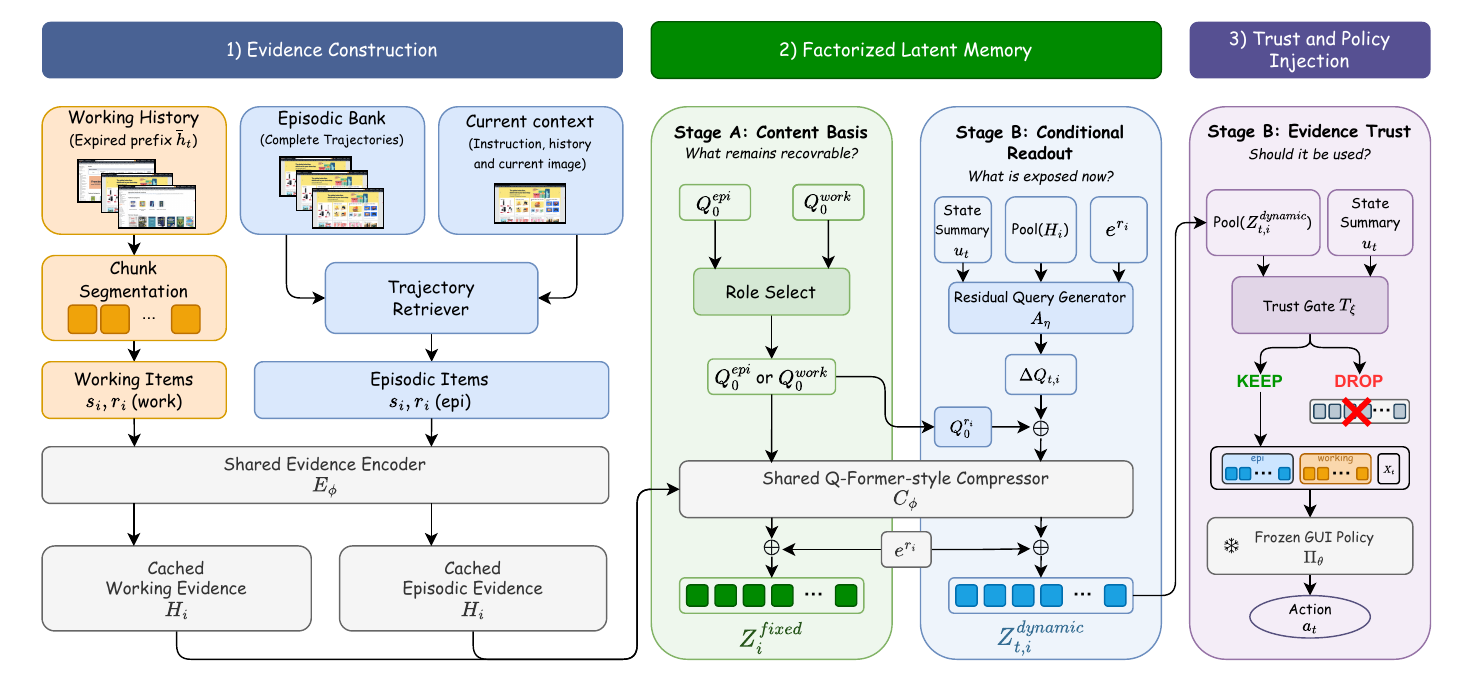}}
\caption{\method retrieves complete episodic trajectories, segments
expired working history, and encodes each evidence item once.
Stage~A learns a role-aware content basis, while Stage~B jointly
optimizes state-conditioned readout and block-level evidence trust.
Accepted dynamic blocks are prepended to the current-context
embeddings for action prediction by the frozen GUI policy.}
\label{fig:method-overview}
\end{figure*}
% !TeX root = ../AnonymousSubmission2027.tex

% !TeX root = ../AnonymousSubmission2027.tex
\section{Method}

% !TeX root = ../AnonymousSubmission2027.tex
% \paragraph{Overview.}
% Figure~\ref{fig:method-overview} summarizes \method. We first define the memory-augmented GUI decision problem
% and construct candidate episodic trajectories and working chunks from raw histories.
% We then introduce the shared latent compressor and its three responsibilities: content basis, conditional readout, and evidence trust.
% Finally, we specify how accepted memory enters the frozen policy and how the two training stages optimize the corresponding pathways.
% !TeX root = ../AnonymousSubmission2027.tex

% !TeX root = ../AnonymousSubmission2027.tex
\subsection{Problem Setup}
\label{sec:problem-formulation}

At step \(t\), a GUI agent receives an instruction \(x\) and
observation \(o_t\), and predicts an action \(a_t\). Let
\(e_t=(o_t,a_t)\) and \(E_{<t}=(e_1,\ldots,e_{t-1})\). We split
the interaction history into an expired prefix and a bounded
suffix directly visible to the policy:
\begin{equation}
    E_{<t}=\bar{h}_t \oplus h_t,
    \qquad
    c_t=(x,h_t,o_t),
\end{equation}
where information in \(\bar{h}_t\) must be supplied through
memory.

Memory evidence comes from completed trajectories
\(\tau_n=(x_n,e_{n,1:T_n})\) in an episodic bank
\(\mathcal{B}^{\mathrm{raw}}\), and from the expired prefix
\(\bar{h}_t\) of the current episode. At step \(t\), these sources
produce evidence sets \(\mathcal{S}^{\mathrm{epi}}_t\) and
\(\mathcal{S}^{\mathrm{work}}_t\). Each evidence item \(s_i\) has a role
\(r_i\in\{\mathrm{epi},\mathrm{work}\}\) and is mapped to a
\(K\)-token latent block:
\begin{equation}
    Z_{t,i}
    =\mathcal{M}_{\Phi}(s_i;c_t,r_i)
    \in\mathbb{R}^{K\times d},
\end{equation}
where \(d\) is the embedding dimension of the GUI policy.
The construction of the two evidence sets is described next.

% !TeX root = ../AnonymousSubmission2027.tex

% !TeX root = ../AnonymousSubmission2027.tex
\subsection{Trajectory Evidence Construction}
\label{sec:evidence}

\paragraph{Retrieved episodic trajectories.}
We use complete trajectories as the basic episodic evidence items.
A fixed trajectory-level retriever encodes each bank trajectory
and the current decision state as
\begin{equation*}
k_n
=
F_{\mathrm{ret}}(\tau_n),
\qquad
q_t
=
F_{\mathrm{ret}}(c_t).
\end{equation*}
The top-$M$ trajectories under cosine similarity form the episodic
evidence set:
\begin{equation*}
R_t
=
\operatorname{TopM}_{n\in[N]}
\cos(q_t,k_n),
\qquad
S_t^{\mathrm{epi}}
=
\{\tau_n\mid n\in R_t\}.
\end{equation*}
Each retrieved trajectory constitutes one evidence item and is later
mapped to one $K$-token latent block. The retriever is not optimized
with the memory pathway, and all matched variants receive the same
retrieved candidates.

% \paragraph{Segmented working evidence.}
% The expired working prefix bypasses retrieval and is partitioned into ordered,
% non-overlapping segments,
% \begin{equation}
%     \mathcal S_t^{\mathrm{work}}
%     =
%     \operatorname{Segment}_{W}(\bar h_t).
% \end{equation}
% Each working chunk is a separate memory item and also yields \(K\) tokens.
% This asymmetric granularity reflects the two lifecycles: completed trajectories
% can be encoded and cached offline, whereas growing working history requires
% incremental chunks. Both item types use the same latent interface. Evidence
% construction determines which items are available; \method determines how each
% item is compressed for the current decision and whether it is used.
\paragraph{Segmented working evidence.}
We construct working-memory evidence directly from the expired
history prefix $\bar{h}_t$. Specifically, $\bar{h}_t$ is partitioned
into ordered, non-overlapping contiguous chunks:
\begin{equation*}
\mathcal{S}^{\mathrm{work}}_t
=
\operatorname{Segment}_{W}(\bar{h}_t),
\end{equation*}
where $W$ is the maximum number of interaction events in each chunk.
Each chunk forms one working-memory evidence item and is assigned the
same $K$-token latent budget as an episodic item.

Together, these standard constructions yield the episodic and working
evidence sets $\mathcal{S}^{\mathrm{epi}}_t$ and
$\mathcal{S}^{\mathrm{work}}_t$; the next section specifies how FocusMem
represents and uses each evidence item.
% !TeX root = ../AnonymousSubmission2027.tex

% !TeX root = ../AnonymousSubmission2027.tex
\subsection{\method{}}
\label{sec:focusmem}

\paragraph{Shared latent compressor.}
Following prior trajectory-based latent memory methods, episodic and
working items share an evidence encoder and a Q-Former-style latent
compressor:
\begin{equation}
H_i=E_\phi(s_i),
\qquad
C_\Phi(s_i;Q)
=
P_\omega\!\left(G_\psi(H_i;Q)\right)
\in\mathbb R^{K\times d}.
\end{equation}
where $Q\in\mathbb{R}^{K\times h}$ contains $K$ query tokens,
$G_{\psi}$ cross-attends from these queries to the encoded trajectory
features $H_i$, and \(P_\omega\) projects the resulting tokens to the frozen
policy dimension $d$. The same encoder and compressor are shared by
both memory roles.

% !TeX root = ../AnonymousSubmission2027.tex
\paragraph{Role-Aware Content Basis: What Remains Recoverable?}
\label{sec:content-basis}

Prior dual-source latent memory shares the same learned queries across
episodic and working evidence, while source embeddings only mark their
origins. This does not explicitly control which role-specific information
survives compression. We therefore introduce role-specific base queries:
\begin{equation}
Z_i^{\mathrm{fixed}}
=
C_{\Phi}(s_i;Q^{r_i}_0)+e^{r_i},
\qquad
r_i\in\{\mathrm{epi},\mathrm{work}\},
\end{equation}
where $Q^{r_i}_0$ controls role-dependent extraction and $e^{r_i}$
identifies the source. The resulting representation is independent of the
current decision state.

For each item, the frozen policy provides two supervision
signals: a generic semantic teacher distribution and a
role-specific functional pseudo-target \(\hat{v}_i^{r_i}\).
Episodic targets emphasize reusable decisions, procedures,
and constraints, whereas working targets emphasize state
changes, progress, and remaining constraints. Under a fixed
semantic prompt, let \(p^{\mathrm{sem}}_{\mathrm{raw},ij}\) and
\(p^{\mathrm{sem}}_{\mathrm{lat},ij}\) denote the frozen policy's
next-token distributions conditioned on the raw evidence item
and \(Z_i^{\mathrm{fixed}}\), respectively.
We optimize
\begin{align}
\mathcal{L}_{\mathrm{basis}}
&=
\lambda_{\mathrm{sem}}\mathcal{L}_{\mathrm{sem}}
+
\lambda_{\mathrm{func}}\mathcal{L}_{\mathrm{func}},\\
\mathcal{L}_{\mathrm{sem}}
&=
\sum_{i,j}
D_{\mathrm{KL}}
\left(
\operatorname{sg}
[p^{\mathrm{sem}}_{\mathrm{raw},ij}]
\middle\|
p^{\mathrm{sem}}_{\mathrm{lat},ij}
\right),\\
\mathcal{L}_{\mathrm{func}}
&=
-\sum_i
\log
\Pi_\theta
\left(
\hat{v}^{r_i}_i
\mid
Z_i^{\mathrm{fixed}},
\rho_{\mathrm{func}}^{r_i}
\right).
\end{align}
The semantic KL transfers broadly verbalizable trajectory content,
whereas functional supervision emphasizes the control-relevant
information required by each memory role. Gradients update only the
memory pathway.
% !TeX root = ../AnonymousSubmission2027.tex

% !TeX root = ../AnonymousSubmission2027.tex
\paragraph{State-Conditioned Readout: What Is Exposed for the Current Decision?}
\label{sec:conditional-readout}

The role-aware content basis encourages each memory item to
retain broadly useful and role-specific information under the
latent bottleneck.
However,
recoverability alone does not determine which part of that information should
influence the current action. A complete episodic trajectory may span search,
filtering, inspection, and verification, while a working chunk may contain both
completed progress and still-active constraints. A fixed representation must
serve all such decision states without knowing which phase is currently relevant.

Let $F_\theta$ denote the frozen policy encoder, and let
\(
u_t=Pool(F_\theta(c_t))
\)
denote the frozen policy representation of the current decision
context. We introduce a lightweight state--item adapter
\(A_\eta\) that generates a residual over the role-specific base
queries:
\begin{equation}
\Delta Q_{t,i}
=
A_\eta\!\left(u_t,Pool(H_i),e^{r_i}\right),
\qquad
Q_{t,i}=Q_0^{r_i}+\Delta Q_{t,i}.
\end{equation}
The resulting decision-facing block is
\begin{equation}
Z^{\mathrm{dynamic}}_{t,i}
=
C_{\Phi}(s_i;Q_{t,i})+e^{r_i}.
\end{equation}
Because the residual depends jointly on the current state and
the encoded item, the same encoded evidence can produce
different decision-facing latent blocks across states. We
zero-initialize the output of $A_{\eta}$, so optimization begins from the
role-aware fixed readout learned by the content basis.
% !TeX root = ../AnonymousSubmission2027.tex

% !TeX root = ../AnonymousSubmission2027.tex
\paragraph{Evidence Trust: Should This Memory Influence the Decision?}
\label{sec:evidence-trust}

% \paragraph{Design.}
% Explicit GUI memories can filter or refresh retrieved records before they reach
% the agent~\citep{zeng2026mementogui,zhu2026hymem}. We give latent memory a
% corresponding post-retrieval ability to abstain. For each generated memory item,
% \begin{equation}
% \label{eq:evidence-trust}
% \begin{aligned}
%     g_{t,i}
%     &=
%     T_\xi\!\left(
%       u_t,\operatorname{Pool}(Z_{t,i}^{\mathrm{dynamic}})
%     \right),\\
%     m_{t,i}
%     &=
%     \mathbb I[\sigma(g_{t,i})>\gamma],
% \end{aligned}
% \end{equation}
% In Eq.~\eqref{eq:evidence-trust}, \(g_{t,i}\) is the trust score,
% \(m_{t,i}\) is the binary admission decision, and \(\gamma\) is fixed on
% validation data. Rejected items are removed from policy attention as complete
% blocks.

% \paragraph{Learning.}
% During training, we sample a straight-through HardConcrete
% mask~\citep{louizos2018l0} and supervise only deliberately injected negative
% items \(\mathcal N_t\):
% \begin{equation*}
%     \mathcal L_{\mathrm{gate}}
%     =
%     -\sum_{i\in\mathcal N_t}
%     \log\!\left(1-\sigma(g_{t,i})\right).
% \end{equation*}
% Ordinary retrieved candidates receive no positive gate labels; action imitation
% determines whether retaining them is useful. A validity mask forces padding or
% an empty item to drop. For \(\mathcal L_{\mathrm{gate}}\), gradients through
% \(\operatorname{Pool}(Z_{t,i}^{\mathrm{dynamic}})\) are stopped, so explicit
% negative labels train admission without reshaping the dynamic representation.

State-conditioned readout determines which part of a memory item is exposed,
but exposed evidence may still be misleading, stale, or incompatible with the
current task. We therefore add a lightweight block-level trust module
$T_{\xi}$ that scores each decision-facing memory block:
\begin{equation}
g_{t,i}
=
T_{\xi}\!\left(
u_t,\operatorname{Pool}(Z^{\mathrm{dynamic}}_{t,i})
\right),
\end{equation}
where $u_t$ is the pooled representation of the current decision state. At
inference, the block is retained only when
\begin{equation}
m_{t,i}
=
\mathbb{I}\!\left[\sigma(g_{t,i})>\gamma\right],
\end{equation}
with threshold $\gamma$ selected on validation data; rejected blocks are
removed from policy attention.

During training, we use a straight-through HardConcrete mask
and inject high-confidence irrelevant episodic items sampled
from unrelated tasks, denoted by \(\mathcal N_t\):
\begin{equation}
\mathcal{L}_{\mathrm{gate}}
=
-\sum_{i\in\mathcal{N}_t}
\log\!\left(1-\sigma(g_{t,i})\right).
\end{equation}
Ordinary retrieved items receive no explicit positive labels;
their utility signal comes from the action objective. Gradients
from \(\mathcal L_{\mathrm{gate}}\) are stopped at
\(Z_{t,i}^{\mathrm{dynamic}}\), so gate supervision learns block
usage without directly reshaping its content.
% !TeX root = ../AnonymousSubmission2027.tex

% !TeX root = ../AnonymousSubmission2027.tex

% !TeX root = ../AnonymousSubmission2027.tex
\begin{table*}[!t]
\centering
{\small
\setlength{\tabcolsep}{5.0pt}
\begin{tabular}{@{}llcccccc@{}}
\toprule
\multirow{2}{*}{Method} &
\multirow{2}{*}{Memory} &
\multicolumn{2}{c}{\textbf{MMInA}} &
\multirow{2}{*}{\textbf{DeepShop}} &
\multirow{2}{*}{\textbf{WebVoyager}} &
\multirow{2}{*}{\shortstack{\textbf{Online-}\\\textbf{Mind2Web}}} &
\multirow{2}{*}{\textbf{Overall}} \\
\cmidrule(lr){3-4}
& & Wiki & Shop & & & & \\
\midrule
\multicolumn{8}{@{}l}{\textbf{General VLMs}} \\
GLM-4.1V-9B-Thinking & No Memory & 39.9 & 46.5 & 12.0 & 17.2 & 14.3 & 26.0 \\
InternVL3.5-14B & No Memory & 46.1 & 48.0 & 33.3 & 10.9 & 9.4 & 29.5 \\
Qwen3-VL-32B & No Memory & 69.8 & 62.5 & 52.0 & 28.4 & 18.0 & 46.2 \\
Qwen3.6-27B & No Memory & 46.1 & 59.5 & 46.7 & 26.8 & 16.9 & 39.2 \\
\midrule
\multicolumn{8}{@{}l}{\textbf{GUI Agents}} \\
Fara-7B & No Memory & 47.4 & 38.5 & 38.0 & 31.1 & 25.2 & 36.0 \\
Molmo-8B & No Memory & 45.1 & 41.5 & 37.3 & 33.6 & 27.8 & 37.1 \\
UI-TARS-1.5-7B & No Memory & 41.2 & 35.0 & 28.7 & 26.6 & 21.8 & 30.7 \\
UI-Venus-1.5-8B & No Memory & 50.6 & 44.5 & 36.0 & 28.1 & 18.4 & 35.5 \\
\midrule
\multicolumn{8}{@{}l}{\textbf{Qwen3-VL-8B Memory Study}} \\
Qwen3-VL-8B & No Memory & 42.2 & 41.0 & 27.3 & 19.6 & 12.4 & 28.5 \\
ReasoningBank & Text Memory & 47.4 & 44.5 & 30.0 & 22.6 & 13.9 & 31.7 \\
Retrieved Full Trajectory & Raw MM & 45.1 & 45.5 & 36.7 & 24.2 & 14.3 & 33.2 \\
CoMEM-style~\citep{wu2025comem} & Episodic latent & 49.0 & 47.5 & 34.7 & 27.4 & 18.0 & 35.3 \\
HyMEM-style~\citep{zhu2026hymem} & Hybrid & 52.9 & 47.0 & 38.0 & 30.3 & 20.7 & 37.8 \\
Mem-W-style~\citep{zhang2026memw} & Dual fixed latent & 60.7 & 54.5 & 40.7 & 33.1 & 23.3 & 42.5 \\
FocusMem-Base (Action-only Fixed) & Dual fixed latent & 54.5 & 45.5 & 30.7 & 22.6 & 15.0 & 33.7 \\
\textbf{\method} & Factorized latent & \textbf{65.3} & \textbf{59.0}  & \textbf{48.7}  & \textbf{39.6}  & \textbf{28.9} & \textbf{48.3} \\
\bottomrule
\end{tabular}}
\caption{Interactive task success rate (SR; $\uparrow$) evaluated by
Gemini-3.1-Pro. Overall is the unweighted mean of the five benchmark
SRs computed before rounding. Bold denotes the best result within the
Qwen3-VL-8B memory study.}
\label{tab:main-results}
\end{table*}
% !TeX root = ../AnonymousSubmission2027.tex

% !TeX root = ../AnonymousSubmission2027.tex
\subsection{Memory Injection and Optimization}
\label{sec:training}

Accepted dynamic blocks are concatenated by source and prepended
to the ordinary decision-context embeddings
\(X_t=F_\theta(c_t)\):
\begin{equation}
U_t=[Z_t^{\mathrm{epi}};Z_t^{\mathrm{work}};X_t],
\qquad
a_t\sim\Pi_\theta(\cdot\mid U_t),
\end{equation}
where rejected blocks are omitted from the input. 
Source embeddings already
identify episodic and working blocks, and the GUI policy $\Pi_\theta$ remains
frozen throughout training.

Optimization proceeds in two stages. Stage A establishes the role-aware
content basis before state-dependent readout and trust are introduced. We
disable the query residual and gate, represent every item with
$Z_i^{\mathrm{fixed}}$, and construct the corresponding input
$U_t^{\mathrm{fixed}}$. The objective is
\begin{equation}
\mathcal L_A
=
-\sum_t\log\Pi_\theta(a_t^\star\mid U_t^{\mathrm{fixed}})
+\lambda_{\mathrm{basis}}\mathcal L_{\mathrm{basis}}.
\end{equation}
Thus, fixed memory is trained both to support the demonstrated action and to
preserve semantic and role-specific functional content.

Stage B initializes from Stage A and enables the state--item query residual
and block-level trust gate. The current state produces
$Z_{t,i}^{\mathrm{dynamic}}$, and only retained blocks enter $U_t$. We optimize
\begin{equation}
\mathcal L_B
=
-\sum_t\log\Pi_\theta(a_t^\star\mid U_t)
+\lambda_m\mathcal L_{\mathrm{gate}}.
\end{equation}
Stage B updates the complete memory pathway, including the shared compressor,
conditional readout, and trust module, while keeping the policy frozen.
Because Stage B also updates the shared compressor, we
evaluate whether content remains recoverable from the resulting
Stage-B fixed blocks.
% !TeX root = ../AnonymousSubmission2027.tex

% !TeX root = ../AnonymousSubmission2027.tex

% !TeX root = ../AnonymousSubmission2027.tex

\section{Experiments}

% !TeX root = ../AnonymousSubmission2027.tex
\subsection{Experimental Setup}
\label{sec:experimental-setup}

\paragraph{Benchmarks and protocol.}
We evaluate interactive task success rate (SR) on fixed accessible
subsets of MMInA-Wikipedia, MMInA-Shopping, DeepShop~\citep{lyu2025deepshop},
WebVoyager, and Online-Mind2Web~\citep{xue2025illusion}
($N=308,200,150,598,266$).
Tasks blocked by CAPTCHAs or human verification are removed once
before evaluation, and the resulting subsets are shared across
methods.
Episodes are limited at 15 steps, with unfinished episodes scored
as failures.
To evaluate task completion, Gemini-3.1-Pro relies on benchmark-specific prompts, which are applied uniformly across all compared methods to ensure consistency.

% \paragraph{Memory construction.}
% The episodic bank combines successful Molmoweb
% trajectories with successful Qwen3.6-27B rollouts collected on
% separately authored tasks that share the underlying websites but
% are disjoint from the evaluation task sets.
% Because these banks contain in-domain demonstrations, comparisons
% with No Memory are treated as system-level references.
% All controlled memory variants share the same bank, frozen
% Qwen3-VL-Embedding-2B retriever, and top-$M=3$ candidates.
% We exclude retrieved memory candidates matching the query by task ID, annotated
% instance, or slot-normalized instruction template.
% Retrieval is performed independently at each step using only the
% current top-$M$ set; it is precomputed for training and online
% with caching for evaluation.
% Each trajectory forms one episodic item, while the expired
% history is partitioned into consecutive $W=4$ event chunks,
% with the final chunk possibly shorter. Under the 15-step
% limit, the latest three events remain directly visible, yielding
% at most three episodic and three working items. With $K=8$
% tokens per item, memory uses at most 48 latent
% tokens before gating.
\paragraph{Memory construction.}
The episodic bank combines successful MolmoWeb trajectories
and Qwen3.6-27B rollouts from separately authored,
website-matched tasks disjoint from evaluation.
Because the bank contains in-domain demonstrations, No Memory
serves only as a system-level reference.
All controlled variants share the bank, frozen
Qwen3-VL-Embedding-2B retriever, and top-$M=3$ candidates.
Candidates matching the query by task ID, annotated instance,
or slot-normalized instruction template are excluded.
Retrieval is performed independently at each step, precomputed
for training, and cached online during evaluation.
Each trajectory forms one episodic item; expired history is split
into consecutive $W=4$-event chunks, with the last possibly shorter.
With the latest three events directly visible under the 15-step
limit, this yields at most three items per source. At $K=8$ tokens
per item, memory contains at most 48 latent tokens before gating.

\paragraph{Training and controls.}
The Qwen3-VL-8B policy~\citep{bai2025qwen3} and UGround-V1-7B grounding model~\citep{gou2024navigating} are
frozen, and only the memory pathway is optimized.
Action supervision is constructed from MolmoWeb~\citep{gupta2026molmoweb} and Qwen3.6-27B
rollouts. We retain successful, high-quality trajectories containing
8--15 interaction steps and exclude any trajectory matching an
episodic-bank or evaluation task by task ID, annotated instance,
or slot-normalized instruction template. This results in
23,438 action-supervision trajectories.
Stages~A and B use learning rates of $1\times10^{-5}$ and
$5\times10^{-6}$, respectively, with global batch size 32;
Stage~B initializes from Stage~A.
We set
$\lambda_{\mathrm{sem}}=\lambda_{\mathrm{func}}=0.5$,
$\lambda_{\mathrm{basis}}=0.25$, and
$\lambda_m=0.05$.
The inference threshold $\gamma=0.3$ is selected on a held-out
validation split disjoint from evaluation.
Matched variants share training data, retrieved candidates,
preprocessing, budgets, grounding, decoding, and evaluation.

\paragraph{Baselines.}
No Memory uses only directly visible context, while ReasoningBank~\citep{ouyang2025reasoningbank} provides text memory and Retrieved
Full Trajectory inserts the retrieved raw multimodal evidence.
CoMEM-style~\citep{wu2025comem}, HyMEM-style~\citep{zhu2026hymem}, and Mem-W-style~\citep{zhang2026memw} are adapted
reimplementations based on released repositories where
available and the corresponding method descriptions otherwise.
The suffix ``-style'' denotes implementation under our unified
harness rather than exact reproduction.
% !TeX root = ../AnonymousSubmission2027.tex
\subsection{Main Results}
\label{sec:overall-performance}

Table~\ref{tab:main-results} reports interactive task success
under the common 15-step protocol. \method achieves the
highest Overall score and consistently outperforms all memory
baselines across the five benchmarks. Our primary comparison
is Action-only Fixed, a fully matched state-independent
dual-source latent-memory baseline. The two methods share the
policy, training data, memory bank, retrieved candidates,
latent-token and action budgets, grounding, decoding, and
evaluator. \method improves over this baseline by +10.8,
+13.5, +18.0, +17.0, and +13.9 points on MMInA-Wiki,
MMInA-Shop, DeepShop, WebVoyager, and Online-Mind2Web,
respectively. The largest gains occur on DeepShop and
WebVoyager, with a further substantial improvement on
Online-Mind2Web. The larger gains on DeepShop and WebVoyager are consistent
with the value of adapting the exposed memory view across
changing decision stages.

As system-level references, \method improves over
Qwen3-VL-8B No Memory by nearly 20 points on average and
outperforms Retrieved Full Trajectory by about 15 points overall, despite
representing each trajectory through a bounded latent interface
rather than inserting its complete multimodal context. It also
surpasses CoMEM-style, HyMEM-style, and Mem-W-style on
every benchmark; compared with the strongest of these,
Mem-W-style, the gains range from +4.5 to +8.0 points.
Together, these results show that preserving broadly useful
content, adapting its exposed view to the current decision, and
limiting unreliable evidence yield a compact yet stronger
memory interface.
% !TeX root = ../AnonymousSubmission2027.tex

% !TeX root = ../AnonymousSubmission2027.tex
\subsection{Component Ablation and Efficiency}
\label{sec:ablation-efficiency}

% \paragraph{Setting.}
% We conduct a cumulative ablation on MMInA Shopping with
% Qwen3-VL-8B. Starting from No Memory, Action-only Fixed adds
% fixed dual-source latent memory; Content Basis adds semantic and
% role-specific functional supervision; Dynamic Readout enables
% state--item-conditioned queries with the gate disabled; and
% Evidence Gate yields the full FocusMem. All variants share the
% retriever, evidence construction, latent and action budgets,
% grounding module, decoding configuration, and evaluator. We report
% SR, mean executed steps, Hit-Max, Time/Task, and Time/Step under
% identical inference settings.

% \paragraph{Results.}
% Table~\ref{tab:ablation} shows monotonic gains as the memory
% pathway is progressively strengthened: SR increases from 41.0 to
% 45.5, 50.5, 54.5, and 59.0, with Action-only Fixed, Content
% Basis, Dynamic Readout, and Evidence Gate contributing
% incremental gains of +4.5, +5.0, +4.0, and +4.5 points, respectively.
% Full FocusMem therefore outperforms No Memory by 18.0 points and
% Action-only Fixed by 13.5 points, supporting the complementary
% benefits of the three factorized components. Meanwhile, mean steps
% decrease from 9.43 to 8.39 and Hit-Max from 31.0\% to 24.0\%.
% Although Time/Step increases by $1.70\times$, shorter trajectories
% limit the Time/Task increase to $1.51\times$, partially offsetting
% the per-step memory overhead.
\paragraph{Setting.}
We conduct a cumulative ablation on MMInA Shopping with
Qwen3-VL-8B. Starting from No Memory, we successively add
fixed dual-source latent memory (Action-only Fixed), the
role-aware Content Basis with semantic and functional
supervision, state-aware Dynamic Readout with the
gate disabled, and the Evidence Gate to obtain full FocusMem.
All variants differ only in the progressively added memory
components. We report SR, mean
executed steps, Hit-Max, Time/Task, and Time/Step under
identical inference settings.

% \begin{table*}[!htbp]
% \centering

% {\small
% \setlength{\tabcolsep}{2.2pt}
% \begin{tabular}{@{}lccccc@{}}
% \toprule
% Setting & SR \(\uparrow\) & \#Steps \(\downarrow\) &
% Hit-Max \(\downarrow\) & Time/Task \(\downarrow\) & Time/Step \(\downarrow\) \\
% \midrule
% No Memory & 41.0 & 9.4 & 31.0 & 72.8 & 7.72 \\
% \midrule
% \(\hookrightarrow\) Action-only Fixed
% & 45.5 {\scriptsize(+4.5)} & 9.0 & 28.5 & 79.5 & 8.82 \\
% \(\hookrightarrow\) + Content Basis
% & 50.5 {\scriptsize(+9.5)} & 9.0 & 29.0 & 80.1 & 8.93 \\
% \(\hookrightarrow\) + Dynamic Readout (Gate Off)
% & 54.5 {\scriptsize(+13.5)} & 8.5 & 25.5 & 105.9 & 12.41 \\
% \(\hookrightarrow\) \textbf{+ Evidence Gate (Full \method)}
% & \textbf{59.0} {\scriptsize(+18.0)} & \textbf{8.4} & \textbf{24.0} & 109.9 & 13.10 \\
% \bottomrule
% \end{tabular}}
% \caption{
% Cumulative component ablation. Parentheses report absolute SR
% gains in percentage points over No Memory. Hit-Max is the
% percentage of tasks reaching the 15-step limit; Time/Task and
% Time/Step are measured in seconds.}
% \label{tab:ablation}
% \end{table*}
\begin{table}[!htbp]
\centering
{\small
\setlength{\tabcolsep}{1mm}
\begin{tabular}{@{}lccccc@{}}
\toprule
Setting
& SR \(\uparrow\)
& Steps \(\downarrow\)
& Hit-Max \(\downarrow\)
& \shortstack{T/Task(s)}
& \shortstack{T/Step(s)} \\
\midrule
No Memory
& 41.0
& 9.4
& 31.0
& 72.8
& 7.72 \\
\midrule
\(\hookrightarrow\) + Fixed
& \shortstack{45.5}
& 9.0
& 28.5
& 79.5
& 8.82 \\

\(\hookrightarrow\) + Content
& \shortstack{50.5}
& 9.0
& 29.0
& 80.1
& 8.93 \\

\(\hookrightarrow\) + Readout
& \shortstack{54.5}
& 8.5
& 25.5
& 105.9
& 12.41 \\

\(\hookrightarrow\) \textbf{+ Gate}
& \shortstack{\textbf{59.0}}
& \textbf{8.4}
& \textbf{24.0}
& 109.9
& 13.10 \\
\bottomrule
\end{tabular}
}
\caption{Cumulative component ablation. Components are added in order. Hit-Max is the 15-step-limit rate.}
\label{tab:ablation}
\end{table}

\paragraph{Results.}
Table~\ref{tab:ablation} shows monotonic SR gains as the
memory pathway is progressively strengthened. Action-only
Fixed first improves No Memory by +4.5 points. The Content
Basis, Dynamic Readout, and Evidence Gate then provide
successive gains of +5.0, +4.0, and +4.5 points, respectively.
Thus, all three factorized components add non-trivial value
beyond fixed action-only compression, rather than the overall
improvement being dominated by a single mechanism. Full
FocusMem outperforms No Memory by 18.0 points and
Action-only Fixed by 13.5 points. It also reduces mean
executed steps from 9.4 to 8.4 and Hit-Max from 31.0\% to
24.0\%. Although Time/Step increases by $1.70\times$,
fewer executed steps limit the Time/Task increase to
$1.51\times$.
% !TeX root = ../AnonymousSubmission2027.tex
\subsection{Content-Basis Probes}
\label{sec:generative-probes}

\paragraph{Setting.}
We isolate the recoverability of the Content Basis after
Stage B on two held-out 200-item WebVoyager subsets: episodic
trajectories (Panel~A) and expired working-memory prefixes
(Panel~B). We compare Action-only, Semantic Only
(semantic KL), Functional Only (role-specific QA), and Both.
All variants use fixed base queries, with Dynamic Readout and
the Evidence Gate disabled, so the frozen policy answers only
from the resulting fixed latent blocks. The same Gemini-3.1-Pro judge, blind to the variant, assigns
normalized 0--100 scores for \textbf{Semantic}
recovery of general trajectory content, \textbf{Aligned}
recovery of role-relevant decisions, outcomes, and state
changes, and \textbf{Transfer} to held-out questions about
subgoals and harmful repetition.

\par\smallskip
\noindent\begin{minipage}{\columnwidth}
\centering
\scriptsize
\setlength{\tabcolsep}{15.0pt}
\renewcommand{\arraystretch}{0.93}
\begin{tabular}{@{}lccc@{}}
\toprule
Variant & Semantic \(\uparrow\) & Aligned \(\uparrow\) & Transfer \(\uparrow\) \\
\midrule
\multicolumn{4}{@{}l}{\textit{Panel A: Episodic trajectories (N=200)}} \\
Action-only & 54.2 & 50.8 & 47.1 \\
Semantic Only & 68.9 & 58.6 & 55.4 \\
Functional Only & 60.7 & 69.4 & 64.8 \\
\textbf{Both} & \textbf{74.6} & \textbf{74.1} & \textbf{70.3} \\
\midrule
\multicolumn{4}{@{}l}{\textit{Panel B: Working-memory prefixes (N=200)}} \\
Action-only & 57.6 & 54.9 & 50.5 \\
Semantic Only & 71.8 & 62.4 & 58.7 \\
Functional Only & 63.1 & 73.8 & 68.9 \\
\textbf{Both} & \textbf{76.9} & \textbf{77.2} & \textbf{73.5} \\
\bottomrule
\end{tabular}
\captionof{table}{Content-basis recoverability after Stage B. Both panels use
200 held-out WebVoyager items, fixed queries, and no dynamic readout or gate.}
\label{tab:generative-probes}
\end{minipage}
\par\smallskip

\paragraph{Result.}
Table~\ref{tab:generative-probes} shows that action supervision
alone yields a weakly recoverable content basis. Semantic Only
primarily improves general trajectory reconstruction, whereas
Functional Only produces larger gains on role-aligned and unseen
transfer queries. Their complementary pattern holds for both
episodic and working memory, and combining the two objectives
achieves the best score on every metric. With Dynamic Readout and the gate disabled, these results
localize the probe improvements to the fixed content basis
and show that both general and role-relevant information
remain recoverable after subsequent optimization.

% !TeX root = ../AnonymousSubmission2027.tex

% !TeX root = ../AnonymousSubmission2027.tex
\subsection{Oracle-Centered Evidence Expansion}
\label{sec:oracle-expansion}

\paragraph{Setting.}
We evaluate 100 WebVoyager decisions that Qwen3-VL-8B
fails under No Memory, focusing on cases where memory can
affect the next action. For each decision, annotators identify
the most relevant fragment from a supporting trajectory,
denoted the \emph{Oracle Core}. We progressively expand it
with \(+2\) and \(+4\) neighboring steps, followed by the
complete trajectory (\emph{Full traj.}).

We train separate Fixed and Dynamic checkpoints for
\(B \in \{4,8,16\}\), where \(B\) denotes the number of output latent
tokens per memory block; all other settings are matched.
Gemini-3.1-Pro judges whether each predicted action is a valid next step
given the task, current state and previous history.

\paragraph{Result.}
Fixed and Dynamic perform nearly identically when given only
the compact, decision-relevant Oracle Core. Their difference
emerges as surrounding trajectory context is added. From the
Oracle Core to the full trajectory, Fixed drops by \(22\),
\(25\), and \(25\) points at \(B=4,8,16\), whereas Dynamic
drops by only \(12\), \(9\), and \(6\) points. Dynamic Readout
therefore provides limited benefit when evidence is already
concise and decision-sufficient, but remains substantially more
robust when relevant content must compete with additional
trajectory stages under a compact bottleneck. Increasing the
latent budget further reduces this degradation for Dynamic,
while Fixed remains sensitive to evidence expansion.

\par\smallskip
\noindent\begin{minipage}{\columnwidth}
\centering
\includegraphics[width=\columnwidth]{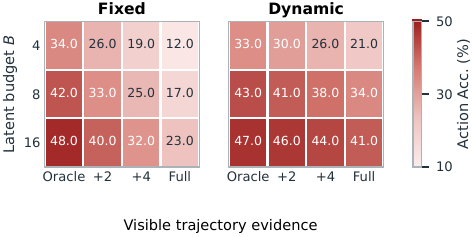}
\captionof{figure}{
Oracle-centered evidence expansion on 100 WebVoyager
decisions. Fixed and Dynamic use identical evidence under
matched latent-token budgets.}
\label{fig:oracle-expansion}
\end{minipage}
\par\smallskip

% !TeX root = ../AnonymousSubmission2027.tex

% !TeX root = ../AnonymousSubmission2027.tex
\begin{figure}[!t]
\centering
\includegraphics[width=\columnwidth]{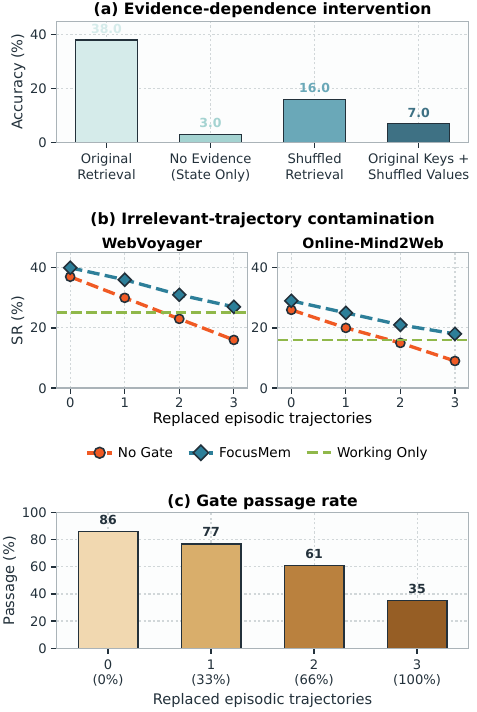}
\captionof{figure}{
Evidence-dependence and episodic-contamination diagnostics.
(a) One-step Action Accuracy under controlled evidence
interventions. (b) Interactive SR as \(0\)--\(3\) retrieved
episodic trajectories are replaced. (c) Gate passage rate.}
\label{fig:evidence-dependence-trust}
\end{figure}

\subsection{Evidence Dependence and Trust}
\label{sec:evidence-dependence-trust}

\paragraph{Setting.}
We conduct two diagnostics at different scales. On the same 100
WebVoyager decisions used above, we disable the gate and fix the
checkpoint, state, and memory slots while comparing
\emph{Original Retrieval}, \emph{No Evidence},
\emph{Shuffled Retrieval}, and \emph{Original Keys + Shuffled
Values}. The same judge scores one-step Action Accuracy, testing
whether Dynamic Readout actually uses retrieved trajectory
values.
Separately, on independent 100-task WebVoyager and
Online-Mind2Web subsets, at every step we replace \(0\)--\(3\) of
the current top-three episodic trajectories with clearly irrelevant
trajectories from unrelated tasks, selected as high-confidence irrelevant negatives. We compare the no-gate
variant, full \method, and Working-Only using interactive SR and
episodic-block passage rate; working memory and unreplaced
trajectories remain unchanged.

\paragraph{Result.}
Panel~(a) shows that Dynamic Readout genuinely relies on
retrieved trajectory content. Original Retrieval performs best;
same-domain shuffling retains only limited procedural benefit,
while removing evidence or breaking key--value correspondence
largely eliminates the gain. Panels~(b) and~(c) show that, as
irrelevant episodic trajectories are injected at every step, the
no-gate variant degrades sharply, whereas full \method remains
9--11 points stronger under maximum contamination. Meanwhile,
the passage rate falls from \(86\%\) to \(35\%\), indicating that
the gate increasingly suppresses contaminated memory blocks.

% !TeX root = ../AnonymousSubmission2027.tex

\section{Limitation}
FocusMem is evaluated primarily with a frozen Qwen3-VL-8B
policy on web benchmarks; transfer to other backbones and
mobile or desktop environments remains open. The trust
diagnostic uses injected irrelevant episodic trajectories and
does not cover subtler mismatches or stale working memory.
All scores rely on Gemini-3.1-Pro, so fixed,
method-blind prompts reduce but do not eliminate judge bias.
% !TeX root = ../AnonymousSubmission2027.tex
\section{Conclusion}

FocusMem factorizes latent GUI memory into content
recoverability, state-conditioned readout, and evidence trust.
Content probes, oracle-centered expansion, and controlled
evidence and contamination diagnostics support the intended
roles of these components, while matched interactive
evaluation shows consistent gains over fixed action-only
compression and prior memory adaptations. These results
suggest that compact latent memory benefits from separating
what is retained, what is exposed, and what is allowed to
influence.
% !TeX root = ../AnonymousSubmission2027.tex

\clearpage      % start the bibliography on a fresh page
{
    \small
    \bibliographystyle{ieeenat_fullname}
    \bibliography{main}
}

% ==================== Merged Supplementary Material ====================
\clearpage
\setcounter{secnumdepth}{2}
% \twocolumn[...] internally, which would override \onecolumn and leave the title
% alone on page 1; so we switch to one column and set the title manually.
\onecolumn
% aaai2027.sty sets \flushbottom, which stretches inter-paragraph glue to fill
% pages that end early (e.g. before a tall pinned figure), producing large gaps.
\raggedbottom
\thispagestyle{plain}
\begin{center}
{\LARGE\bfseries FocusMem: Factorizing Content, Readout, and Trust in Latent GUI Memory}\\[0.4em]
{\LARGE\bfseries Supplementary Material}
\end{center}
\vspace{1.2em}
\appendix

\noindent{\large\bfseries Contents}\par
\vspace{2pt}
\begingroup
\small
\setlength{\parskip}{0pt}
\setlength{\baselineskip}{10pt}
\pdfcontentsentry{toc-a}{\textbf{A\quad Detailed Algorithmic Design}}
  {\pageref{app:algorithmic-design}}
\pdfcontentssubentry{toc-a1}{A.1\quad Evidence Encoding and Fixed Readout}
  {\pageref{app:evidence-encoding}}
\pdfcontentssubentry{toc-a2}{A.2\quad Stage A: Role-Aware Content Formation}
  {\pageref{app:stage-a}}
\pdfcontentssubentry{toc-a3}{A.3\quad Stage B: State-Conditioned Readout and Evidence Trust}
  {\pageref{app:stage-b}}
\pdfcontentssubentry{toc-a4}{A.4\quad Trainable Pathway Configurations}
  {\pageref{app:trainable-pathway}}
\pdfcontentsentry{toc-b}{\textbf{B\quad Prompt and Input Assembly}}
  {\pageref{app:prompt-assembly}}
\pdfcontentssubentry{toc-b1}{B.1\quad Raw Evidence Serialization}
  {\pageref{app:raw-evidence}}
\pdfcontentssubentry{toc-b2}{B.2\quad Semantic Distribution Prompt}
  {\pageref{app:semantic-prompt}}
\pdfcontentssubentry{toc-b3}{B.3\quad Teacher--Student Separation}
  {\pageref{app:teacher-student}}
\pdfcontentsentry{toc-c}{\textbf{C\quad Data, Memory Bank, and Splits}}
  {\pageref{app:data}}
\pdfcontentssubentry{toc-c1}{C.1\quad Evaluation Benchmarks and Accessible Subsets}
  {\pageref{app:benchmarks}}
\pdfcontentssubentry{toc-c2}{C.2\quad Trajectory Data Construction}
  {\pageref{app:construction}}
\pdfcontentssubentry{toc-c3}{C.3\quad High-Quality Selection}
  {\pageref{app:selection}}
\pdfcontentssubentry{toc-c4}{C.4\quad Dataset Splits}
  {\pageref{app:splits}}
\pdfcontentssubentry{toc-c5}{C.5\quad Leakage Prevention}
  {\pageref{app:leakage}}
\pdfcontentssubentry{toc-c6}{C.6\quad Data Representation}
  {\pageref{app:representation}}
\pdfcontentsentry{toc-d}{\textbf{D\quad Evaluation and Annotation Protocols}}
  {\pageref{app:eval}}
\pdfcontentssubentry{toc-d1}{D.1\quad Metrics}
  {\pageref{app:metrics}}
\pdfcontentssubentry{toc-d2}{D.2\quad Automated Judges}
  {\pageref{app:judges}}
\pdfcontentssubentry{toc-d3}{D.3\quad Diagnostic Construction and Configuration}
  {\pageref{app:diagconfig}}
\pdfcontentssubentry{toc-d4}{D.4\quad Run Protocol}
  {\pageref{app:runs}}
\pdfcontentssubentry{toc-d5}{D.5\quad Human Annotation Protocol}
  {\pageref{app:annotation}}
\pdfcontentsentry{toc-e}{\textbf{E\quad Software and Hardware Environment}}
  {\pageref{app:environment}}
\pdfcontentsentry{toc-f}{\textbf{F\quad Training Details and Hyperparameter Search}}
  {\pageref{app:training-details}}
\pdfcontentssubentry{toc-f1}{F.1\quad Training Details}
  {\pageref{app:training-spec}}
\pdfcontentssubentry{toc-f2}{F.2\quad Hyperparameter Search}
  {\pageref{app:hyperparameters}}
\pdfcontentsentry{toc-g}{\textbf{G\quad Qualitative Case Studies}}
  {\pageref{app:cases}}
\pdfcontentssubentry{toc-g1}{G.1\quad Content Recoverability (Episodic Reusable Experience)}
  {\pageref{app:case1}}
\pdfcontentssubentry{toc-g2}{G.2\quad Working Memory Retains Progress}
  {\pageref{app:case2}}
\pdfcontentssubentry{toc-g3}{G.3\quad State-Conditioned Readout (Fixed vs.\ Dynamic)}
  {\pageref{app:case3}}
\endgroup
\vspace{1ex}

\pdfcontentsdest{toc-a}
\section{Detailed Algorithmic Design}
\label{app:algorithmic-design}

\pdfcontentsdest{toc-a1}
\subsection{Evidence Encoding and Fixed Readout}
\label{app:evidence-encoding}

For an evidence item \(s_i\), let
\(H_i=E_\phi(s_i)\in\mathbb{R}^{L_i\times d_e}\) be its
multimodal trajectory features and \(b_i\in\{0,1\}^{L_i}\) the
valid-token mask. Episodic items contain one complete retrieved trajectory;
working items contain one chronological chunk of expired interaction history.
The shared compressor first maps the evidence features to its hidden space,
\(\widetilde H_i=\operatorname{LN}(H_i)W_{\mathrm{in}}\), and initializes
its latent sequence with the role-specific queries
\(L_i^{(0)}=Q_0^{r_i}\in\mathbb{R}^{K\times h}\). Each Q-Former layer applies
masked cross-attention followed by a feed-forward block:
\begin{align}
\widehat L_i^{(\ell)}
&=L_i^{(\ell-1)}
 +\operatorname{XAttn}
 \left(L_i^{(\ell-1)},\widetilde H_i,\widetilde H_i;b_i\right),\\
L_i^{(\ell)}
&=\widehat L_i^{(\ell)}
 +\operatorname{FFN}\!\left(\widehat L_i^{(\ell)}\right).
\end{align}
The fixed policy-facing block is
\begin{equation}
Z_i^{\mathrm{fixed}}
=\operatorname{LN}\!\left(L_i^{(D)}\right)W_{\mathrm{out}}+e^{r_i}
\in\mathbb{R}^{K\times d}.
\end{equation}
The mask prevents padding from affecting cross-attention, while the additive
role embedding marks every token in the resulting block.

\pdfcontentsdest{toc-a2}
\subsection{Stage A: Role-Aware Content Formation}
\label{app:stage-a}

Let \(\mathcal{R}(s_i)\) denote the raw multimodal serialization of an item.
Under a generic description prompt \(\rho_{\mathrm{sem}}\), the frozen policy
first generates a teacher description \(\hat{\ell}_i\). At token position \(j\),
the raw and latent teacher-forced distributions are
\begin{align}
p_{\mathrm{raw},ij}^{\mathrm{sem}}
&=\Pi_\theta\!\left(\cdot\mid
\mathcal{R}(s_i),\rho_{\mathrm{sem}},\hat{\ell}_{i,<j}\right),\\
p_{\mathrm{lat},ij}^{\mathrm{sem}}
&=\Pi_\theta\!\left(\cdot\mid
Z_i^{\mathrm{fixed}},\rho_{\mathrm{sem}},\hat{\ell}_{i,<j}\right).
\end{align}
Using the same prefix on both sides isolates information lost by the latent
bottleneck. For functional supervision, the raw path produces a role-specific
target \(\hat v_i^{r_i}\), and the latent path reconstructs that target:
\begin{align}
\mathcal L_{\mathrm{sem}}
&=\sum_{i,j}D_{\mathrm{KL}}\!\left(
\sg[p_{\mathrm{raw},ij}^{\mathrm{sem}}]
\middle\|p_{\mathrm{lat},ij}^{\mathrm{sem}}\right),\\
\mathcal L_{\mathrm{func}}
&=-\sum_{i,j}\log\Pi_\theta\!\left(
\hat v_{i,j}^{r_i}\mid
Z_i^{\mathrm{fixed}},\rho_{\mathrm{func}}^{r_i},
\hat v_{i,<j}^{r_i}\right).
\end{align}
For a decision example, all fixed episodic and working blocks are ordered by
source and evidence rank to form \(U_t^{\mathrm{fixed}}\). The token-normalized
action loss uses \(N_{\mathrm{act}}=\sum_t|a_t^\star|\):
\begin{equation}
\mathcal L_{\mathrm{act}}^{\mathrm{fixed}}
=-\frac{1}{N_{\mathrm{act}}}\sum_{t,j}\log\Pi_\theta\!\left(
a_{t,j}^{\star}\mid U_t^{\mathrm{fixed}},a_{t,<j}^{\star}\right).
\end{equation}

\begin{focusalgorithm}{Stage A --- Role-Aware Content Formation}
\label{alg:stage-a}
\begin{tabularx}{\linewidth}{@{}r@{\hspace{0.7em}}X@{}}
\textbf{Input} &
Decision demonstrations \(\mathcal D_{\mathrm{act}}\), raw evidence items
\(\mathcal D_{\mathrm{mem}}\), frozen GUI policy \(\Pi_\theta\), prompts
\(\rho_{\mathrm{sem}}\) and \(\rho_{\mathrm{func}}^r\).\\[2pt]
1 & Initialize
\(\Phi\equiv\{\phi,\psi,\omega,Q_0^{\mathrm{epi}},Q_0^{\mathrm{work}},
e^{\mathrm{epi}},e^{\mathrm{work}}\}\);
\(\operatorname{Freeze}(\Pi_\theta)\), \(\Delta Q_{t,i}\leftarrow0\),
\(m_{t,i}\leftarrow1\).\\
2 & \(\mathcal L_{\mathrm{sem}},\mathcal L_{\mathrm{func}},
\mathcal L_{\mathrm{act}}^{\mathrm{fixed}}\leftarrow0\).\\
3 & \textbf{for} \(s_i\in\mathcal B_{\mathrm{mem}}\subset
\mathcal D_{\mathrm{mem}}\) \textbf{do}\\
4 & \quad \(H_i\leftarrow E_\phi(s_i)\), \quad
\(Z_i^{\mathrm{fixed}}\leftarrow
C_\Phi(s_i;Q_0^{r_i})+e^{r_i}\).\\
5 & \quad \(\hat\ell_i\leftarrow
\operatorname{Decode}_{\Pi_\theta}
(\mathcal R(s_i),\rho_{\mathrm{sem}})\).\\
6 & \quad \(p_{\mathrm{raw},ij}^{\mathrm{sem}}\leftarrow
\Pi_\theta(\cdot\mid\mathcal R(s_i),\rho_{\mathrm{sem}},
\hat\ell_{i,<j})\), \quad
\(p_{\mathrm{lat},ij}^{\mathrm{sem}}\leftarrow
\Pi_\theta(\cdot\mid Z_i^{\mathrm{fixed}},\rho_{\mathrm{sem}},
\hat\ell_{i,<j})\).\\
7 & \quad \(\mathcal L_{\mathrm{sem}}\mathrel{+}=
\sum_jD_{\mathrm{KL}}\!\left(
\sg[p_{\mathrm{raw},ij}^{\mathrm{sem}}]\middle\|
p_{\mathrm{lat},ij}^{\mathrm{sem}}\right)\).\\
8 & \quad \(\hat v_i^{r_i}\leftarrow
\operatorname{Decode}_{\Pi_\theta}
(\mathcal R(s_i),\rho_{\mathrm{func}}^{r_i})\).\\
9 & \quad \(\mathcal L_{\mathrm{func}}\mathrel{-}=
\sum_j\log\Pi_\theta(\hat v_{i,j}^{r_i}\mid
Z_i^{\mathrm{fixed}},\rho_{\mathrm{func}}^{r_i},
\hat v_{i,<j}^{r_i})\).\\
10 & \textbf{end for}\\
11 & \textbf{for} \((c_t,a_t^\star)\in\mathcal B_{\mathrm{act}}
\subset\mathcal D_{\mathrm{act}}\) \textbf{do}\\
12 & \quad \(\mathcal I_t\leftarrow
\operatorname{Retrieve}_{\mathrm{epi}}(c_t)
\cup\operatorname{Segment}_{\mathrm{work}}(c_{<t})\).\\
13 & \quad \(\mathcal Z_t^{\mathrm{fixed}}\leftarrow
\{C_\Phi(s_i;Q_0^{r_i})+e^{r_i}:i\in\mathcal I_t\}\).\\
14 & \quad \(U_t^{\mathrm{fixed}}\leftarrow
[\operatorname{Order}_{\mathrm{src,rank}}
(\mathcal Z_t^{\mathrm{fixed}});F_\theta(c_t)]\).\\
15 & \quad \(\mathcal L_{\mathrm{act}}^{\mathrm{fixed}}
\mathrel{-}=\frac{1}{N_{\mathrm{act}}}\sum_j
\log\Pi_\theta(a_{t,j}^\star\mid
U_t^{\mathrm{fixed}},a_{t,<j}^\star)\).\\
16 & \textbf{end for}\\
17 & \(\mathcal L_{\mathrm{basis}}\leftarrow
\lambda_{\mathrm{sem}}\mathcal L_{\mathrm{sem}}
+\lambda_{\mathrm{func}}\mathcal L_{\mathrm{func}}\), \quad
\(\mathcal L_A\leftarrow\mathcal L_{\mathrm{act}}^{\mathrm{fixed}}
+\lambda_{\mathrm{basis}}\mathcal L_{\mathrm{basis}}\).\\
18 & \(\Phi\leftarrow
\operatorname{OptStep}(\Phi,\nabla_\Phi\mathcal L_A)\).\\
\textbf{Output} & Stage-A memory parameters \(\Phi_A\).\\
\end{tabularx}
\end{focusalgorithm}

\pdfcontentsdest{toc-a3}
\subsection{Stage B: State-Conditioned Readout and Evidence Trust}
\label{app:stage-b}

The frozen policy encodes the decision context as
\(u_t=\pool(F_\theta(c_t))\). An item summary
\(\bar h_i=\sum_{\ell}b_{i\ell}H_{i\ell}/\sum_{\ell}b_{i\ell}\)
is combined with the decision state and role:
\begin{align}
d_{t,i}
&=\operatorname{LN}(W_u u_t+W_h\bar h_i+W_r e^{r_i}),\\
\Delta Q_{t,i}
&=\operatorname{reshape}_{K\times h}
\left(W_2\,\operatorname{GELU}(W_1d_{t,i}+b_1)+b_2\right),\\
Q_{t,i}&=Q_0^{r_i}+\Delta Q_{t,i}.
\end{align}
Zero-initializing \(W_2,b_2\) makes the initial dynamic readout identical to
Stage A. Re-running the shared compressor with \(Q_{t,i}\) produces
\(Z_{t,i}^{\mathrm{dynamic}}\).

The trust module then makes one admission decision for each complete latent
block. It compares the decision state with the pooled dynamic block:
\begin{align}
g_{t,i}
&=T_\xi\!\left(
u_t,\pool(Z_{t,i}^{\mathrm{dynamic}})\right),\\
m_{t,i}
&=\mathbb I[\sigma(g_{t,i})>\gamma]
\qquad\text{at inference},
\end{align}
where \(\gamma\) is selected on validation data. During training, a
straight-through HardConcrete sample replaces the deterministic mask so action
gradients can determine whether retaining an ordinary candidate is useful.
Deliberately injected negative episodic items provide the explicit drop loss
\begin{equation}
\mathcal L_{\mathrm{gate}}
=-\sum_t\sum_{i\in\mathcal N_t}
\log\!\left(1-\sigma(g_{t,i})\right).
\end{equation}
The gate-loss gradient is stopped at
\(Z_{t,i}^{\mathrm{dynamic}}\); padding and empty items are always rejected.
Blocks zeroed by the mask are discarded during ordering and never enter the
policy input.
\begin{focusalgorithm}{Stage B --- State-Conditioned Readout and Evidence Trust}
\label{alg:stage-b}
\begin{tabularx}{\linewidth}{@{}r@{\hspace{0.7em}}X@{}}
\textbf{Input} &
Stage-A parameters \(\Phi_A\), decision demonstrations
\(\mathcal D_{\mathrm{act}}\), decision-paired negative episodic items
\(\mathcal N_t\), frozen policy \(\Pi_\theta\).\\[2pt]
1 & \(\Phi\leftarrow\Phi_A\), \quad
\(\eta,\xi\leftarrow\operatorname{Init}()\), \quad
\((W_2,b_2)\leftarrow(0,0)\), \quad
\(\operatorname{Freeze}(\Pi_\theta)\).\\
2 & \(\mathcal L_{\mathrm{act}}^{\mathrm{dyn}},
\mathcal L_{\mathrm{gate}}\leftarrow0\).\\
3 & \textbf{for} \((c_t,a_t^\star,\mathcal N_t)
\in\mathcal B_{\mathrm{act}}\) \textbf{do}\\
4 & \quad \(\mathcal I_t\leftarrow
\operatorname{Retrieve}_{\mathrm{epi}}(c_t)
\cup\operatorname{Segment}_{\mathrm{work}}(c_{<t})
\cup\mathcal N_t\), \quad
\(u_t\leftarrow\pool(F_\theta(c_t))\).\\
5 & \quad \textbf{for} \(i\in\mathcal I_t\) \textbf{do}\\
6 & \qquad \(H_i\leftarrow E_\phi(s_i)\), \quad
\(\bar h_i\leftarrow
\sum_\ell b_{i\ell}H_{i\ell}/\sum_\ell b_{i\ell}\).\\
7 & \qquad \(d_{t,i}\leftarrow
\operatorname{LN}(W_uu_t+W_h\bar h_i+W_re^{r_i})\).\\
8 & \qquad \(\Delta Q_{t,i}\leftarrow
\operatorname{reshape}_{K\times h}
(W_2\operatorname{GELU}(W_1d_{t,i}+b_1)+b_2)\).\\
9 & \qquad \(Q_{t,i}\leftarrow Q_0^{r_i}+\Delta Q_{t,i}\), \quad
\(Z_{t,i}^{\mathrm{dynamic}}\leftarrow
C_\Phi(s_i;Q_{t,i})+e^{r_i}\).\\
10 & \qquad \(g_{t,i}\leftarrow
T_\xi(u_t,\pool(Z_{t,i}^{\mathrm{dynamic}}))\), \quad
\(\widetilde m_{t,i}\leftarrow
\mathbb I[\lVert b_i\rVert_1>0]\,
\operatorname{HC}_{\mathrm{ST}}(g_{t,i})\).\\
11 & \quad \textbf{end for}\\
12 & \quad \(\mathcal B_t\leftarrow
\{\widetilde m_{t,i}\odot Z_{t,i}^{\mathrm{dynamic}}:i\in\mathcal I_t\}\).\\
13 & \quad \(U_t\leftarrow
[\operatorname{Order}_{\mathrm{src,rank}}(\mathcal B_t);
F_\theta(c_t)]\).\\
14 & \quad \(\mathcal L_{\mathrm{act}}^{\mathrm{dyn}}
\mathrel{-}=\frac{1}{N_{\mathrm{act}}}\sum_j
\log\Pi_\theta(a_{t,j}^\star\mid U_t,a_{t,<j}^\star)\).\\
15 & \quad \(\mathcal L_{\mathrm{gate}}\mathrel{-}=
\sum_{i\in\mathcal N_t}\log\!\left(
1-\sigma\!\left[T_\xi\!\left(
u_t,\pool(\sg[Z_{t,i}^{\mathrm{dynamic}}])\right)\right]\right)\).\\
16 & \textbf{end for}\\
17 & \(\mathcal L_B\leftarrow
\mathcal L_{\mathrm{act}}^{\mathrm{dyn}}
+\lambda_m\mathcal L_{\mathrm{gate}}\), \quad
\((\Phi,\eta,\xi)\leftarrow
\operatorname{OptStep}((\Phi,\eta,\xi),
\nabla_{\Phi,\eta,\xi}\mathcal L_B)\).\\
\textbf{Output} & \(\Phi_B\equiv(\Phi,\eta,\xi)\) and
validation-selected threshold \(\gamma\).\\
\end{tabularx}
\end{focusalgorithm}

\pdfcontentsdest{toc-a4}
\subsection{Trainable Pathway Configurations}
\label{app:trainable-pathway}

Table~\ref{tab:trainable-pathway-configurations} details the trainable
compression pathway. LoRA
is applied only to the compression backbone; the independent policy copy
remains frozen and carries no adapters.

\begin{table*}[!t]
\centering
\small
\setlength{\tabcolsep}{5pt}
\begin{tabularx}{\textwidth}{@{}p{0.17\textwidth}p{0.10\textwidth}X
>{\raggedleft\arraybackslash}p{0.20\textwidth}@{}}
\toprule
Compression backbone & Module & Structure & Trainable parameters\\
\midrule
Qwen3-VL-8B \(H=4096\)
& LoRA &
\(r=64,\ \alpha=16,\ p=0.05\); 36 text blocks, targeting
\(\{q,k,v,o,\mathrm{gate},\mathrm{up},\mathrm{down}\}\) projections
(252 linear modules) &
174,587,904 (174.59M)\\
\cmidrule(lr){2-4}
& Q-Former &
\(K=8\), 16 heads (\(d_h=256\)), eight shared refinement steps,
\(4H\) FFN, \(H{\rightarrow}H\) input/output projections, two role vectors,
and \(H{\rightarrow}256{\rightarrow}8H\) dynamic-query adapter &
234,995,712 core \(+\) 9,478,400 adapter \(=\) 244,474,112 (244.47M)\\
\bottomrule
\end{tabularx}
\caption{Trainable compression-path configurations. LoRA excludes the vision
tower, token embeddings, language-model head, Q-Former, and frozen policy
copy. Q-Former counts include learned queries, input/output projections, the
shared attention--FFN block, role vectors, and the dynamic-query adapter.
Trust-gate parameters are excluded.}
\label{tab:trainable-pathway-configurations}
\end{table*}
% \FloatBarrier

\pdfcontentsdest{toc-b}
\section{Prompt and Input Assembly}
\label{app:prompt-assembly}

\pdfcontentsdest{toc-b1}
\subsection{Raw Evidence Serialization}
\label{app:raw-evidence}

Each memory item is serialized as a sequence of multimodal user turns in
chronological order. A header specifies the current task when applicable, the
memory role, and the number of steps. Each subsequent turn contains one
screenshot followed by its recorded action. For episodic memory, the complete
retrieved successful trajectory is one item; for working memory, each
chronological chunk is one item. Teacher prompts consume this raw
serialization, whereas student prompts replace it with the corresponding
\(K\)-token latent block.

\pdfcontentsdest{toc-b2}
\subsection{Semantic Distribution Prompt}
\label{app:semantic-prompt}

Semantic supervision uses a short, role-neutral instruction asking the frozen
policy to describe the supplied GUI trajectory using only visible screenshots
and recorded actions. The raw path first produces the description
\(\hat\ell_i\). Raw and latent paths are then evaluated with the same instruction
and the same teacher-forced prefix \(\hat\ell_{i,<j}\); the training target is
the full raw next-token distribution rather than a generated hard label. Since
this branch changes only the conditioning source, no separate long-form prompt
template is required.

Functional supervision retains two memory roles but instantiates them with four
complementary QA subtypes: high-level experience and anchor recovery for
episodic memory, and latest-state and observed-change recovery for working
memory. All four contribute to the single role-aware
\(\mathcal L_{\mathrm{func}}\) objective.

\begin{promptbox}{Functional Supervision Prompt Assembly}
\small
\begin{minipage}[t]{0.48\textwidth}
\textbf{Raw teacher}

\medskip
\textbf{[SYSTEM]}\\
You create concise supervision for a \(K\)-token GUI memory compressor.
Use only the supplied chronological screenshots and recorded actions.
Return valid JSON with exactly the requested fields and no additional prose.

\medskip
\textbf{[USER]}\\
\slot{FUNCTIONAL\_REQUEST}\\
Return exactly: \slot{OUTPUT\_SCHEMA}\\
Current task: \slot{SOURCE\_TASK}\\
Memory trajectory:

\slot{SCREENSHOT\_1}\\
Recorded action: \slot{ACTION\_1}\\
\(\cdots\)\\
\slot{SCREENSHOT\_L}\\
Recorded action: \slot{ACTION\_L}

\medskip
\textbf{[ASSISTANT]}\\
\slot{TEACHER\_FUNCTIONAL\_TARGET}
\end{minipage}
\hfill
\begin{minipage}[t]{0.48\textwidth}
\textbf{Latent student}

\medskip
\textbf{[CONTINUOUS PREFIX]}\\
\slot{FIXED\_MEMORY\_BLOCK \(Z_i^{\mathrm{fixed}}\)}

\medskip
\textbf{[SYSTEM]}\\
You decode a GUI-agent memory represented by \(K\) latent tokens.
Return one concise valid JSON object with exactly the requested fields.
Do not use markdown or add prose.

\medskip
\textbf{[USER]}\\
\slot{FUNCTIONAL\_REQUEST}\\
Return exactly: \slot{OUTPUT\_SCHEMA}

\medskip
\textbf{[ASSISTANT TARGET]}\\
\slot{TEACHER\_FUNCTIONAL\_TARGET}
\end{minipage}

\medskip
\hrule
\medskip
\textbf{For each subtype, insert the text into
\slot{FUNCTIONAL\_REQUEST} and \slot{OUTPUT\_SCHEMA}, respectively.}

\begin{tabularx}{\linewidth}{@{}
>{\raggedright\arraybackslash\bfseries}p{0.16\linewidth}
>{\raggedright\arraybackslash}X
>{\raggedright\arraybackslash}X@{}}
\toprule
Subtype & Text inserted into \slot{FUNCTIONAL\_REQUEST} &
Text inserted into \slot{OUTPUT\_SCHEMA}\\
\midrule
Episodic summary &
Extract compact, task-relevant, transferable guidance: progress, reusable
experience, task constraints, and the anchor decision. Do not narrate every
step. &
\texttt{\{"progress\_summary":"...", "useful\_experience":"...",
"task\_constraints":"...", "anchor\_decision":"..."\}}\\
Episodic anchor &
Recover only the final anchor state, its successful action, and the visible
evidence supporting that decision. &
\texttt{\{"anchor\_state":"...", "anchor\_action":\{"name":"...",
"arguments":\{...\}\}, "decision\_basis":"..."\}}\\
Working state &
Recover the latest observed state, important visible facts, current progress,
remaining constraints, and the last recorded action. Do not claim that the
last action succeeded without a later screenshot. &
\texttt{\{"latest\_observed\_state":"...", "important\_state\_facts":"...",
"current\_progress":"...", "remaining\_constraints":"...",
"last\_recorded\_action":\{...\}\}}\\
Working change &
Summarize only state changes visible across the chronological screenshots; do
not infer the unobserved result of the final action. &
\texttt{\{"initial\_state":"...", "observed\_change":"...",
"latest\_observed\_state":"..."\}}\\
\bottomrule
\end{tabularx}
\end{promptbox}
\captionof{figure}{Prompt assembly for functional supervision. The raw teacher produces
the supervision target from the original trajectory, and the latent student
reconstructs the same target from fixed memory tokens.}
\label{fig:functional-prompt-template}

\begin{promptbox}{Action Imitation Prompt Assembly}
\small
\begin{minipage}[t]{0.48\textwidth}
\textbf{[CONTINUOUS PREFIX]}\\
\slot{EPISODIC\_BLOCK\_1}\(\ldots\)\\
\slot{EPISODIC\_BLOCK\_M'}\\
\slot{WORKING\_BLOCK\_1}\(\ldots\)\\
\slot{WORKING\_BLOCK\_J'}

\medskip
\textbf{[SYSTEM]}\\
You are a GUI agent that completes tasks by interacting with a web page
through screenshots. Decide the single best next action and return exactly one
JSON action object.

Available actions:\\
\slot{GROUNDING\_BACKEND\_TOOL\_SCHEMA}

Track what the task needs, what has already been completed, and what the
current screenshot shows. Do not repeat an action that failed to change the
page. Stop as soon as the requested answer is visible, and ground the answer
only in observed page content.

\medskip
\textbf{[USER: RECENT HISTORY]}\\
\texttt{\# Recent history (oldest first)}\\
\slot{HISTORY\_SCREENSHOT\_1}\\
\texttt{Step \slot{STEP\_ID\_1}: \slot{RECORDED\_ACTION\_1}}\\
\(\cdots\)\\
\slot{HISTORY\_SCREENSHOT\_H}\\
\texttt{Step \slot{STEP\_ID\_H}: \slot{RECORDED\_ACTION\_H}}
\end{minipage}
\hfill
\begin{minipage}[t]{0.48\textwidth}
\textbf{[USER: TASK]}\\
\texttt{**Current task:** \slot{TASK\_INSTRUCTION}}

\medskip
\textbf{[USER: ACTION BUDGET]}\\
\texttt{ACTION NUMBER LEFT: You have \slot{ACTIONS\_LEFT} actions left.
Finish the task within the remaining actions. If only 1 action is left, you
MUST output the stop action with your answer.}

\medskip
\textbf{[USER: CURRENT OBSERVATION]}\\
\slot{CURRENT\_SCREENSHOT}\\
\texttt{This is the current screenshot.}

\medskip
\textbf{[USER: FINAL REMINDER]}\\
\texttt{Do not repeat a failed action. If you can already answer, stop now.
Answer from what the page shows. Output exactly one JSON action object and no
prose outside the JSON.}

\medskip
\textbf{[ASSISTANT TARGET / OUTPUT]}\\
\texttt{\{"name":"\slot{ACTION\_NAME}",
"arguments":\slot{ACTION\_ARGUMENTS}\}}
\end{minipage}
\end{promptbox}
\captionof{figure}{Prompt assembly for action imitation. Continuous episodic and working
memory blocks precede the ordinary GUI-agent messages, while only the
demonstrated assistant action contributes to the imitation loss.}
\label{fig:action-prompt-template}

\pdfcontentsdest{toc-b3}
\subsection{Teacher--Student Separation}
\label{app:teacher-student}

All pseudo-target generation uses the raw evidence serialization and a frozen
teacher policy. Memory-path optimization uses the latent student condition and
never backpropagates into the teacher. For action imitation, prompt tokens are
masked from the loss and only the demonstrated assistant action contributes
token-level cross-entropy. Functional targets are similarly teacher-forced,
while semantic supervision retains the complete raw-policy distribution at
every description position.

% ---- Added by jjc: finalized supplement content (Data & Splits, Evaluation) ----
% ============================================================
% Appendix section: Data, Memory Bank, Splits, Leakage Prevention
% Added by jjc from the finalized supplement draft (§5). Case studies excluded.
% Uses collaborator conventions: \method, wideprompt, booktabs. Verbatim needs fancyvrb.
% ============================================================
\pdfcontentsdest{toc-c}
\section{Data, Memory Bank, Splits, and Leakage Prevention}
\label{app:data}

This section details the data underlying \method. Two trajectory corpora feed
the memory pathway: (i) an \emph{action-supervision set} used to train the
role-aware content basis, conditional readout, and trust gate over the frozen
policy (Stages~A--B), and (ii) an \emph{episodic memory bank} that serves as the
retrieval corpus at both training and evaluation time. Working-memory evidence
is not stored in a bank; it is constructed online from the expired prefix of the
current episode. We describe the evaluation benchmarks and their accessible
subsets (Table~\ref{tab:benchmarks}), the episodic bank and action-supervision
corpus together with their filtering
(\S\ref{app:construction}--\S\ref{app:selection}), and the splits used across the
main and diagnostic experiments (Table~\ref{tab:splits}). Because the memory
pathway is trained and evaluated with retrieved evidence, preventing overlap
between the memory sources and the evaluation tasks is essential:
\S\ref{app:leakage} specifies the three-level matching---by task identifier, by
annotated instance, and by slot-normalized instruction template---used both to
exclude retrieved candidates matching the current query and to remove training
trajectories matching any evaluation task. Throughout, we report corpus
statistics and representative, de-identified examples; full task-identifier lists
are omitted to preserve anonymity.

\pdfcontentsdest{toc-c1}
\subsection{Evaluation Benchmarks and Accessible Subsets}
\label{app:benchmarks}

\begin{center}
\small
\captionof{table}{Evaluation benchmarks: number of tasks in the fixed subset we evaluate,
and the success judge for each benchmark (full judge prompts in
\S\ref{app:eval}). All judges are instantiated with the same LLM used throughout
the paper (Gemini-3.1-Pro) under fixed, benchmark-specific prompts. Across all
benchmarks, the judge is invoked only on trajectories that emit an explicit final
answer (a \texttt{stop} action) within the 15-step budget; a trajectory that
reaches the step cap without producing an answer is scored as a failure without a
judge call.}
\label{tab:benchmarks}
\begin{tabularx}{\textwidth}{@{}l l r X@{}}
\toprule
Benchmark & Environment & \#Tasks & Success judge (brief)\\
\midrule
MMInA-Wikipedia & Kiwix mirror (offline) & 308 & MMInA official LLM \emph{fuzzy-match} of the agent's final answer against the task's gold answer\\
MMInA-Shopping & OneStopMarket (self-hosted) & 200 & same MMInA official LLM fuzzy-match against the gold answer\\
DeepShop & live web & 150 & DeepShop official judge: an LLM checks whether the final answer / selected product satisfies the request's constraints\\
WebVoyager & live web & 598 & WebVoyager official judge: an LLM inspects the final answer together with the trajectory's last screenshots to decide completion\\
Online-Mind2Web & live web & 266 & Online-Mind2Web official \emph{WebJudge}: a multi-stage LLM procedure (task key points $\rightarrow$ key screenshots $\rightarrow$ action/observation history)\\
\bottomrule
\end{tabularx}
\end{center}

The two MMInA suites run against local or self-hosted environments---a Kiwix
Wikipedia mirror and the OneStopMarket shopping site---so every official task is
reachable, and we evaluate on their full official subsets. The three live-web
benchmarks (DeepShop, WebVoyager, Online-Mind2Web) additionally contain tasks
that can fail for reasons outside the agent's control---the site being
unreachable, login walls, CAPTCHAs, WAF or geographic blocks, ad-hijack
redirects, or site-pushed modal dialogs. To obtain a single reproducible subset
per benchmark, we screen each one once before the main experiments. We run the
full benchmark with a reference policy and, for every failed trajectory, an LLM
auditor inspects the \emph{entire} trajectory---the instruction and, for each
step, the action taken followed by the resulting screenshot---and decides whether
the failure was caused by an external, site-imposed obstacle (one of seven
categories: unreachable site, login wall, CAPTCHA, WAF block, geographic block,
site-pushed modal, or ad hijack) rather than by the agent itself.
Whole-trajectory context is essential: it lets the auditor separate a modal the
site pushed unprompted from a dialog the agent's own action triggered (e.g., an
add-to-cart confirmation), which a per-screenshot check confounds. Tasks flagged
as externally obstructed are removed once; the remaining subset is frozen and
shared identically across all methods and runs. This yields the final counts in
Table~\ref{tab:benchmarks} (e.g., WebVoyager $643\rightarrow598$,
Online-Mind2Web $300\rightarrow266$). The auditor uses the judge LLM at
temperature~0, sees at most 12 evenly sampled steps per trajectory (always
including the first and last), and returns a single OK / BLOCK decision; its
prompt is shown in Figs.~\ref{fig:screen-a},~\ref{fig:screen-b}.

\begin{promptbox}{Task-accessibility screening prompt}
\begin{Verbatim}[fontsize=\small,breaklines=true,breakanywhere=true]
You are auditing a web agent's COMPLETE trajectory to decide one thing:
was the agent's execution obstructed by an EXTERNAL factor - something the website
imposed - as opposed to the agent simply being bad at the task?

You get the task instruction, then every step as: the action the agent took,
followed by the screenshot of the page AFTER that action.

Report BLOCK only for obstacles the SITE imposed on the agent:
  UNREACHABLE - the site failed to load: browser error page (ERR_*, DNS/cert), the
    site's own error page, a blank page, or a stuck loading state with no content.
  LOGIN_WALL  - the content needed for the task sits behind a sign-in the agent
    cannot pass, and the site offered no guest path.
  CAPTCHA     - a human-verification challenge (Turnstile, reCAPTCHA, 'Verify you
    are human', 'Press & Hold', 'Checking your browser').
  WAF_BLOCK   - access denied / request blocked / 403 / 'unusual traffic detected' /
    'access temporarily restricted'.
  GEO_BLOCK   - 'not available in your country/region'.
  AD_HIJACK   - the agent clicked something inside the target site and an AD carried
    the browser OFF to an unrelated third-party domain, derailing the run. Judge this
    by the URL: if it leaves the task's site for an unrelated advertiser domain right
    after a click on what looked like site navigation, that is AD_HIJACK.
  POPUP_BLOCK - a modal the SITE pushed at the agent unprompted (newsletter or email
    capture, discount offer, app-install / QR dialog, cookie-consent modal, survey,
    age gate) that sat on top of the content and derailed the run.
\end{Verbatim}
\end{promptbox}
\captionof{figure}{Task-accessibility screening prompt (\S\ref{app:benchmarks}), part~1:
the auditor's task and the seven site-imposed BLOCK categories. The
not-obstruction rules and output format continue in Fig.~\ref{fig:screen-b}.}
\label{fig:screen-a}

\begin{promptbox}{Task-accessibility screening prompt (continued)}
\begin{Verbatim}[fontsize=\small,breaklines=true,breakanywhere=true]
CRITICAL - these are NOT external obstruction. Answer OK for them:
  * A dialog that appeared BECAUSE OF THE AGENT'S OWN ACTION. Look at the action that
    preceded it. Clicking 'Add to cart' -> a cart confirmation dialog. Clicking a
    search box -> a search overlay. Clicking 'Filter' -> a filter panel. Clicking a
    product -> a quick-view. These are the site working correctly.
  * A site-pushed popup that the agent DISMISSED or simply worked around, and the run
    continued normally afterwards - it did not obstruct execution.
  * Thin banner strips along an edge (cookie bar, site notice) that leave the content
    usable.
  * The agent clicking the wrong thing, wandering, looping, giving up, running out of
    steps, or answering incorrectly. That is agent capability, NOT obstruction.
  * A page that simply does not contain the answer, or empty search results.
  * A login page the agent navigated to BY ITS OWN CHOICE when the task did not
    require logging in.

Decide on the trajectory as a whole: did an external obstacle actually STOP the
agent from proceeding?

Answer on ONE line, exactly:
  OK
or
  BLOCK <KIND> step<N> <short reason quoting the on-screen text>
When unsure, answer OK.
\end{Verbatim}
\end{promptbox}
\captionof{figure}{Task-accessibility screening prompt (\S\ref{app:benchmarks}), part~2:
the not-obstruction (OK) rules and output format. The auditor receives
\texttt{TASK: <intent>} followed by each step as the action taken and the
resulting screenshot ($>$12 steps are evenly subsampled, always keeping the first
and last), runs at temperature~0, and returns a single \texttt{OK} or
\texttt{BLOCK <KIND> step<N> <reason>} line over the seven fixed categories.}
\label{fig:screen-b}

\pdfcontentsdest{toc-c2}
\subsection{Trajectory Data Construction}
\label{app:construction}

The training set and the episodic bank are drawn from successful, in-domain
trajectories that share the evaluation websites but are disjoint from the
evaluation tasks (\S\ref{app:leakage}). We build them by two routes.

\paragraph{Rewrite route (WebVoyager, Online-Mind2Web).}
For the two domains covered by WebVoyager and Online-Mind2Web, we adapt
trajectories from the open MolmoWeb corpus, whose tasks are seeded from these
benchmarks. We first quality-filter the MolmoWeb trajectories with (i)
programmatic rules and (ii) an LLM evaluator, and then rewrite the survivors: an
LLM regenerates the per-step, first-person reasoning into a concise \emph{state}
description conditioned only on the task, the prior action history, the current
screenshot, and the action actually taken (which is correct, since the run
succeeded), with an explicit constraint never to reference any external hint.
Only the reasoning text is regenerated; the actions and screenshots are
unchanged. Quality filtering reduces the pool from $17{,}474$ to $7{,}864$
trajectories for WebVoyager and from $18{,}329$ to $7{,}803$ for
Online-Mind2Web before rewriting. The rewrite prompt is shown in
Fig.~\ref{fig:rw}.

\paragraph{Generate-and-rollout route (MMInA-Shopping, MMInA-Wikipedia, DeepShop).}
For these domains we synthesize fresh trajectories. We first generate task seeds.
For the shopping domains we scrape real product names from the store catalog,
sample a small set of products together with a task type (product-detail QA,
price, cheapest/most-expensive, or a two-product comparison), and prompt an LLM
to write a self-contained task grounded in those exact products; the
MMInA-Shopping seed prompt is shown in Fig.~\ref{fig:seedshop}, and DeepShop
seeds are produced by the same product-grounded procedure on the DeepShop store.
For MMInA-Wikipedia we prompt an LLM to write self-contained single-article
factual or visual questions directly (Fig.~\ref{fig:seedwiki}). Each seed is a
(\texttt{start\_url}, \texttt{task}) pair (\S\ref{app:representation}). We then
execute every seed with Qwen3.6-27B and keep only the trajectories that a strict
vision judge (temperature~0) admits as successful, requiring the final answer to
be grounded in a screenshot and every step to be sound (correct target, no
degenerate loops, ending in an explicit \texttt{stop}).

\begin{promptbox}{Reasoning-rewrite prompt (rewrite route)}
\begin{Verbatim}[fontsize=\small,breaklines=true,breakanywhere=true]
You reconstruct the step-by-step reasoning of a SUCCESSFUL web agent, to use as training
data. For one step you get: the overall TASK, the agent's prior ACTION HISTORY, the CURRENT
screenshot, and the ACTION the agent took at this step (this action is correct - the run
succeeded). Write the FIRST-PERSON reasoning the agent would have written to justify THIS
action from THIS state.
Rules:
- First person, present tense, as the agent deciding its next action in real time.
- CRITICAL: reason ONLY from the task, your own memory of the prior steps, and the current
  screenshot. NEVER mention or allude to a note, hint, demonstrator, label, or any
  information 'given'/'provided' to you, and never say the action is correct/given or that
  this is reconstructed. Write exactly as the agent would, with no meta-references.
- Ground everything in what is actually visible in the CURRENT screenshot; name concrete
  on-screen elements (buttons, fields, links, text).
- The reasoning must justify EXACTLY the given action (same target element / same typed
  text); do not propose a different action.
- State only what the screenshot + history support; do not invent page content.
Return one concise string (2-4 sentences) covering (1) what the task needs, (2) what you
have done/learned so far from the history, (3) what the current screenshot shows; then why
this action is the next step.
\end{Verbatim}
\end{promptbox}
\captionof{figure}{Reasoning-rewrite prompt used in the rewrite route. Only the per-step
reasoning is regenerated; the actions and screenshots are unchanged. The prompt
produces a concise \emph{state} description and forbids any reference to external
hints.}
\label{fig:rw}

\begin{promptbox}{MMInA-Shopping task-seed generation prompt}
\begin{Verbatim}[fontsize=\small,breaklines=true,breakanywhere=true]
You generate NEW evaluation-style tasks for a shopping web agent operating on the 'One Stop
Market' online store. You are given a list of REAL products currently in the store's catalog.
Each task MUST be about one (or, for a comparison, two) of the EXACT products given - use the
product's real name so the agent can find it. Write a short, self-contained question
answerable by browsing that product page or a small comparison. Allowed types: product-detail
QA (color/size/material/count/weight/whether it includes or comes with X), price,
cheapest/most-expensive variant, compare two of the given products, rating/review lookup.
Style like: 'What color is the X?', 'Does the X come with a Y?', 'Which is heavier, X or Y?',
'How many pieces are in the X?'. Do NOT invent products that are not in the given list. Return
ONLY a JSON array; each item = {"question": "...", "answer": "<short answer: a word, number,
color, or Yes/No>", "product": "<the exact product name(s) from the list this is about>"}.
\end{Verbatim}
\end{promptbox}
\captionof{figure}{MMInA-Shopping task-seed generation prompt (product-grounded). The user
message supplies the sampled real products. DeepShop seeds use the same
scrape-products-plus-task-type procedure on its own catalog.}
\label{fig:seedshop}

\begin{promptbox}{MMInA-Wikipedia task-seed generation prompt}
\begin{Verbatim}[fontsize=\small,breaklines=true,breakanywhere=true]
You generate NEW evaluation-style questions for a Wikipedia-reading web agent. Each question
must be a short, self-contained factual or visual question answerable from a SINGLE English
Wikipedia article (geography, landmarks, history, science, biology, people, culture). Style:
like 'Is the Sydney Opera House by the sea?', 'Are there cliffs in Three Natural Bridges?',
'From the map, is Nanning coastal?'. Prefer yes/no or one-phrase answers. Diversify entities
and topics; do NOT reuse the example entities. Return ONLY a JSON array; each item =
{"question": "...", "answer": "<short answer, e.g. Yes / No / a word or number>", "entity":
"<the Wikipedia page title it is about>"}.
\end{Verbatim}
\end{promptbox}
\captionof{figure}{MMInA-Wikipedia task-seed generation prompt: self-contained
single-article factual or visual questions.}
\label{fig:seedwiki}

\pdfcontentsdest{toc-c3}
\subsection{High-Quality Selection}
\label{app:selection}

Pooling the successful trajectories from both routes, we retain a high-quality
subset in three stages: (i) a rule layer that removes unparseable actions,
episodes not ending in an explicit \texttt{stop}, adjacent-duplicate actions, and
other degenerate patterns; (ii) an LLM rubric that scores each surviving
trajectory for correctness, step necessity, reasoning coherence, monotone
trajectory shape, and termination discipline, applying hard caps that reject any
trajectory with a severe defect; and (iii) a length filter keeping trajectories
of $8$--$15$ steps. This yields the $23{,}438$ high-quality training trajectories
used in the main paper. The rubric prompt is shown in Fig.~\ref{fig:rubric}.
Per-domain counts and the training-vs-bank split are given in
Table~\ref{tab:splits}.

\begin{promptbox}{High-quality scoring rubric}
\begin{Verbatim}[fontsize=\small,breaklines=true,breakanywhere=true]
You are auditing a GUI-agent trajectory that will be used as TRAINING data. Your job is to
decide whether the whole episode reads like a competent human who already knows this website
doing the task - deliberate, minimal, and correct - or like a model fumbling toward an answer.
Be harsh. "Nothing obviously wrong" is a 6, not a 9. When in doubt, score lower.

{MODE_NOTE}   # process: cannot see screenshots, judge process only; full: screenshots given.

# What you are given
- TASK: the instruction the agent was given.
- STEPS: the ordered actions. Each has the action name, its arguments, and the agent's own
  `reasoning` for that step.{IMAGE_NOTE}

# Dimensions (judge each, then produce ONE overall score)
A. Correctness - every part of the instruction satisfied; final `stop` answer right.
B. Necessity - every step load-bearing; delete-test each step; exploratory clicks / useless
   scrolls / re-reading = waste.
C. Coherence - reasoning refers to the actual current situation and justifies THIS action;
   generic filler across steps is a red flag.
D. Trajectory shape - monotone progress; backtracking, re-searching after finding,
   back-and-forth scrolling, undo steps = serious defects.
E. Termination discipline - stop at the FIRST step the answer was obtainable; answer is the
   shortest phrase (a full sentence is a defect).

# Hard caps (overall score is the MINIMUM of the caps that match)
- Final answer wrong/unsupported .......................... cap 2
- Task only partially completed ........................... cap 3
- Any recovery / backtrack / re-search after finding ...... cap 4
- Generic-filler reasoning on 2+ steps, or contradicts action .. cap 5
- Any deletable step ...................................... cap 6
- Stopped later than possible ............................. cap 6
- Answer is a sentence/paragraph, not shortest phrase ..... cap 7
- Trivially short for what the task demanded .............. cap 8

# Score anchors: 10 flawless (rare) . 8 clean but unremarkable . 6 correct with process waste
# . 4 correct after a wrong turn . 3 partial . 2 wrong . 0-1 incoherent.

# Output - ONLY a JSON object:
{"score": <0-10>, "A_correct": <0-10>, "B_necessary": <0-10>, "C_coherent": <0-10>,
 "D_shape": <0-10>, "E_termination": <0-10>,
 "caps_applied": ["<hard caps applied, [] if none>"],
 "worst_step": <0-based index of worst step, or -1>,
 "why": "<one sentence, max 30 words>"}
\end{Verbatim}
\end{promptbox}
\captionof{figure}{High-quality scoring rubric. \texttt{\{MODE\_NOTE\}} and
\texttt{\{IMAGE\_NOTE\}} are instantiated for the process pass (actions plus
reasoning only) and the full pass (screenshots included for final verification).}
\label{fig:rubric}

\pdfcontentsdest{toc-c4}
\subsection{Dataset Splits}
\label{app:splits}

Table~\ref{tab:splits} summarizes the purposes and disjointness
relationships of the principal data splits. Unreported sizes are marked
with a dash.

\begin{center}
\small
\captionof{table}{Dataset splits. Every split is
disjoint from the evaluation tasks under the three-level match of
\S\ref{app:leakage}.}
\label{tab:splits}
% TODO(authors): fill Size and per-domain breakdown; values intentionally deferred.
\begin{tabularx}{\textwidth}{@{}l X l l@{}}
\toprule
Split & Purpose & Size & Disjoint from\\
\midrule
Action-supervision train & Train Stages A--B (content basis, readout, gate) & 23{,}438 & episodic bank, all eval tasks\\
Episodic bank & Retrieval corpus (train + eval time) & 25{,}737 & all eval tasks\\
Eval: MMInA-Wikipedia & Main results & 308 & train, bank\\
Eval: MMInA-Shopping & Main results & 200 & train, bank\\
Eval: DeepShop & Main results & 150 & train, bank\\
Eval: WebVoyager & Main results & 598 & train, bank\\
Eval: Online-Mind2Web & Main results & 266 & train, bank\\
Contamination & WebVoyager + Online-Mind2Web subsets & $100+100$ & --- \\
\bottomrule
\end{tabularx}
\end{center}

\pdfcontentsdest{toc-c5}
\subsection{Leakage Prevention}
\label{app:leakage}

Because the memory pathway is trained and evaluated with retrieved evidence, we
prevent overlap between the memory sources and the evaluation tasks using the
same three-level match---by \emph{task identifier}, by \emph{annotated instance},
and by \emph{slot-normalized instruction template}---at two points. First, when
constructing the training set, we remove any trajectory that matches an
\emph{episodic-bank} or \emph{evaluation} task under any of the three keys.
Second, at both training and evaluation time we exclude from the retrieved
candidate set any trajectory that matches the current query under any of the
three keys, so a query never retrieves a memory of itself or of a task-equivalent
instance.

\pdfcontentsdest{toc-c6}
\subsection{Data Representation}
\label{app:representation}

A task seed is a (\texttt{start\_url}, \texttt{task}) pair: the URL of the page
where the episode starts and the natural-language instruction. A trajectory is
the task together with, for each step, the screenshot the agent observed and the
action it took. Each action is serialized as a JSON object (Fig.~\ref{fig:action-schema}),
where only the arguments relevant to the action type are present. Screenshots are
stored as PNG at the policy's input resolution; URLs, account identifiers, and
any personal information are removed. We omit a full example trajectory here to
save space.

\begin{promptbox}{Action serialization schema}
\begin{Verbatim}[fontsize=\small,breaklines=true,breakanywhere=true]
{"name": "<click | type | scroll | press_key | goto | wait | stop>",
 "arguments": {"description": "<target element (click / type)>",
               "text":        "<typed text (type)>",
               "direction":   "<up | down (scroll)>",
               "key":         "<key (press_key)>",
               "url":         "<url (goto)>",
               "answer":      "<final answer (stop)>",
               "reasoning":   "<first-person state / justification>"}}
\end{Verbatim}
\end{promptbox}
\captionof{figure}{Action serialization schema. Only the arguments relevant to the action
type are present in any given step.}
\label{fig:action-schema}

% ============================================================
% Appendix section: Evaluation and Annotation Protocols
% Added by jjc from the finalized supplement draft (§8). Case studies excluded.
% ============================================================
\pdfcontentsdest{toc-d}
\section{Evaluation and Annotation Protocols}
\label{app:eval}

This section defines the reported metrics, task-completion and diagnostic
judges, per-diagnostic construction and controls, run and parsing protocols,
and the human annotations used in the oracle-centered and contamination
experiments. All automated judgments use Gemini-3.1-Pro at temperature~0
with fixed prompts applied identically across compared variants. Judge inputs
contain no method, checkpoint, or intervention identifier.

\pdfcontentsdest{toc-d1}
\subsection{Metrics}
\label{app:metrics}

Table~\ref{tab:metrics} defines every metric we report and indicates where each is
used in the main paper. Task success and one-step action validity are determined by the corresponding
judges. Content-probe scores are assigned by the diagnostic content judge.
Hit-Max, executed steps, wall-clock time, and gate passage are computed
directly from rollout and gate logs.

\begin{center}
\small
\captionof{table}{Metrics. All success judgments use Gemini-3.1-Pro
(\S\ref{app:judges}). Table and figure references in the last column point to the
main paper.}
\label{tab:metrics}
\begin{tabularx}{\textwidth}{@{}l X l@{}}
\toprule
Metric & Definition & Used in\\
\midrule
SR (task success rate) & Fraction of tasks the judge marks completed under the benchmark protocol within the 15-step cap. A task with no explicit \texttt{stop} answer is a failure scored without a judge call (\S\ref{app:benchmarks}). & Tables~1,~2; Fig.~4(b)\\
Hit-Max & Fraction of tasks that reach the 15-step limit. & Table~2\\
\#Steps & Mean number of executed steps per task. & Table~2\\
Time/Task, Time/Step & Mean wall-clock seconds per task and per executed step. & Table~2\\
Semantic & Mean normalized judge score for role-neutral probes about the task
goal, salient visual evidence, constraints, recorded actions, and temporal
state represented by the fixed latent block. & Table~3\\
Aligned & Mean normalized judge score for held-out role-aligned probes:
reusable procedure and anchor decision for episodic memory, and latest state
and observed change for working memory. & Table~3\\
Transfer & Mean normalized judge score for held-out questions about remaining
subgoals and harmful action repetition, which are not used as content-basis
training targets. & Table~3\\
Action Acc. & Fraction of decisions for which the predicted action is judged
to be a valid next step given the task, recent history, current screenshot,
and action interface. Multiple valid actions may receive credit; exact match
to a single reference action is not required. & Figs.~3,~4(a)\\
Gate passage rate
& Fraction of all non-empty episodic candidate blocks admitted by the trust gate under a given contamination level. We micro-average over all executed steps, candidate slots, and tasks from the WebVoyager and Online-Mind2Web contamination subsets.
& Fig.~4(c)\\
Overall (aggregation) & Unweighted mean of the five per-benchmark SRs, computed before rounding. & Table~1\\
\bottomrule
\end{tabularx}
\end{center}

\paragraph{Content-probe aggregation.}
We use six fixed probe templates in total, one for each
memory-role--probe-category pair:
Semantic-Episodic, Semantic-Working, Aligned-Episodic,
Aligned-Working, Transfer-Episodic, and Transfer-Working.
Each episodic item is evaluated once with each of the three episodic
templates, and each working-memory item is evaluated once with each of
the three working-memory templates. The judge directly returns a score
from $0$ to $100$; no further normalization is applied.

Let
$r\in\{\mathrm{epi},\mathrm{work}\}$ denote the memory role,
$c\in\{\mathrm{sem},\mathrm{align},\mathrm{trans}\}$ the probe category,
and $s_{i,c}^{r}\in[0,100]$ the judge score for item $i$. We report
\begin{equation}
    \operatorname{ProbeScore}_{r,c}
    =
    \frac{1}{N_r}
    \sum_{i=1}^{N_r}
    s_{i,c}^{r},
    \qquad
    N_{\mathrm{epi}}=N_{\mathrm{work}}=200.
    \label{eq:probe-aggregation}
\end{equation}
The three categories are evaluated on the same fixed item manifest
within each role. The episodic and working-memory panels use separately
sampled manifests and are reported independently.

\pdfcontentsdest{toc-d2}
\subsection{Automated Judges}
\label{app:judges}
\subsubsection{Task-Completion Judges}
Task-completion judgments produce SR per benchmark, all with Gemini-3.1-Pro at
temperature~0 and fixed prompts shown in Figs.~\ref{fig:judge-mmina}--\ref{fig:judge-m2w-b}. MMInA-Wikipedia and
MMInA-Shopping compare the agent's final answer to the task's gold answer by LLM
fuzzy matching from text alone (Fig.~\ref{fig:judge-mmina}). DeepShop and
WebVoyager share one judge that assesses the final answer together with the last
five trajectory screenshots (Fig.~\ref{fig:judge-wv}). Online-Mind2Web uses the
official multi-stage WebJudge, which extracts the task's key points, scores every
screenshot for relevance, and issues a final verdict from the key points, the
high-scoring screenshots (relevance~$\ge 3$, capped at fifty), and the action
history; it does not read the agent's final answer (Figs.~\ref{fig:judge-m2w-a},~\ref{fig:judge-m2w-b}).
Across all task-completion judges we apply the pre-filter of
\S\ref{app:benchmarks}: a trajectory that does not emit an explicit \texttt{stop}
answer within the 15-step budget is scored a failure without any judge call, and
tasks removed by the accessibility screen never enter a denominator. All judges
are method-blind: prompts carry no variant, model, or file identifier.

\begin{promptbox}{MMInA fuzzy-match judge (Wikipedia + Shopping; temperature~0; text only)}
\begin{Verbatim}[fontsize=\small,breaklines=true,breakanywhere=true]
[SYSTEM]
You are a helpful evaluator that provides binary yes/no responses.

[USER]
You are an evaluator that determines if a predicted answer is correct for a given question.

Question: {question}
Reference Answer: {reference}
Predicted Answer: {pred}

Please evaluate if the predicted answer is correct. Consider:
1. Semantic similarity to the reference answer
2. Key information overlap
3. Factual accuracy
Note: If the predicted answer is like "I'm sorry, I can't answer that question.", you should
return "no".

Respond with only "yes" or "no":
- "yes": if the predicted answer is correct or equivalent to the reference answer
- "no": if the predicted answer is incorrect, incomplete, or irrelevant
\end{Verbatim}
\end{promptbox}
\captionof{figure}{MMInA fuzzy-match success judge, used for MMInA-Wikipedia and
MMInA-Shopping. Text only (no screenshots); the gold answer comes from the task.}
\label{fig:judge-mmina}

\begin{promptbox}{WebVoyager / DeepShop judge (temperature~0; last 5 screenshots)}
\begin{Verbatim}[fontsize=\small,breaklines=true,breakanywhere=true]
[SYSTEM]
As an evaluator, you will be presented with three primary components to assist you in your
role:

1. Web Task Instruction: a clear, specific directive in natural language detailing the online
activity (search, verify, compare, etc.).

2. Result Screenshots: the last few screens showing the result or intermediate state of the
web task.

3. Result Response: a textual answer produced at the end of browsing.

-- You SHOULD NOT make assumptions based on information not present in the screenshots.
-- Assess the task instruction against the outcome in the screenshots and the response.
-- If the instruction has multiple sub-tasks, failing any one is unsuccessful.
-- The screenshots are authentic; when the response contradicts them, the screenshots prevail;
   when the response adds detail not shown, you may believe it.

Explain briefly, then give a verdict as 'SUCCESS' or 'NOT SUCCESS' between <result> and
</result>.

[USER]
TASK: {intent}
Result Response: {answer}
(followed by the last 5 screenshots)
\end{Verbatim}
\end{promptbox}
\captionof{figure}{WebVoyager / DeepShop success judge (shared). The final answer is
assessed together with the last five trajectory screenshots.}
\label{fig:judge-wv}

\begin{promptbox}{Online-Mind2Web official WebJudge (3-stage; temperature~0)}
\textbf{Stage 1 --- key-point extraction (System)}
\begin{Verbatim}[fontsize=\small,breaklines=true,breakanywhere=true]
You are an expert tasked with analyzing a given task to identify the key points explicitly
stated in the task description.

**Objective**: Carefully analyze the task description and extract the critical elements
explicitly mentioned in the task for achieving its goal.

**Instructions**:
1. Read the task description carefully.
2. Identify and extract **key points** directly stated in the task description.
   - A **key point** is a critical element, condition, or step explicitly mentioned in the
     task description.
   - Do not infer or add any unstated elements.
   - Words such as "best," "highest," "cheapest," "latest," "most recent," "lowest,"
     "closest," "highest-rated," "largest," and "newest" must go through the sort
     function(e.g., the key point should be "Filter by highest").

**Respond with**:
- **Key Points**: A numbered list of the explicit key points for completing this task, one per
  line, without explanations or additional details.
\end{Verbatim}
\textbf{Stage 2 --- per-screenshot scoring (System), one image per call}
\begin{Verbatim}[fontsize=\small,breaklines=true,breakanywhere=true]
You are an expert evaluator tasked with determining whether an image contains information
about the necessary steps to complete a task.

**Objective**: Analyze the provided image and decide if it shows essential steps or evidence
required for completing the task. Use your reasoning to explain your decision before assigning
a score.

**Instructions**:
1. Provide a detailed description of the image, including its contents, visible elements, text
   (if any), and any notable features.
2. Carefully examine the image and evaluate whether it contains necessary steps or evidence
   crucial to task completion.
3. Provide your response in the format:
- **Reasoning**: [explanation, citing specific elements in the image]
- **Score**: 1 (no relevant info) ... 5 (clearly displays necessary steps/evidence).
\end{Verbatim}
\end{promptbox}
\captionof{figure}{Online-Mind2Web official WebJudge, stages~1--2 (key-point extraction and
per-screenshot scoring), reproduced verbatim from OSU-NLP-Group/Online-Mind2Web
(Stage~2's scoring rubric condensed for space). The final-verdict stage is shown
in Fig.~\ref{fig:judge-m2w-b}.}
\label{fig:judge-m2w-a}

\begin{promptbox}{Online-Mind2Web official WebJudge (stage 3 --- final verdict; temperature~0)}
\textbf{Stage 3 --- final verdict (System)}
\begin{Verbatim}[fontsize=\small,breaklines=true,breakanywhere=true]
You are an expert in evaluating the performance of a web navigation agent. Given the user's
task, the agent's action history, key points for task completion, some potentially important
web pages in the agent's trajectory and their reasons, determine whether the agent has
completed the task and achieved all requirements.

Your response must strictly follow the evaluation criteria (filters must be applied and
visibly effective; sort-type key points such as "best/cheapest/latest" must use the sort
function; numeric/range filters must match exactly; some tasks require a submission or a
visible result; empty-but-correctly-performed retrieval still counts as success; if all items
are already shown, an explicit filter is unnecessary).

Format your response into two lines:
Thoughts: <reasoning that double-checks each key point against the criteria>
Status: "success" or "failure"
\end{Verbatim}
\end{promptbox}
\captionof{figure}{Online-Mind2Web official WebJudge, stage~3 (final verdict), reproduced
verbatim from OSU-NLP-Group/Online-Mind2Web (failure-case examples condensed for
space). The judge scores from key points, high-scoring screenshots, and the
action history; it does not read the agent's final answer.}
\label{fig:judge-m2w-b}

\subsubsection{Content-Basis Probe Generation}
\label{app:probe-generation}
For every probe, the policy receives exactly one fixed latent memory block,
the memory role, and the probe instruction. Dynamic Readout and the trust gate
are disabled. The policy does not receive the raw trajectory, the current
decision state, or any teacher-generated target.
\begin{promptbox}{Content-basis probe generation prompt}
\begin{Verbatim}[fontsize=\small,breaklines=true,breakanywhere=true]
[CONTINUOUS PREFIX]
{{FIXED_MEMORY_BLOCK}}

[SYSTEM]
You are decoding a GUI-agent memory that is provided ONLY as the latent tokens in the
continuous prefix above. Answer the question below about what that memory encodes.
- Use ONLY the latent memory and the stated memory role. You are NOT given the raw
  screenshots, the raw action list, the current decision state, or any reference answer.
- Report only what the memory supports; do not invent pages, actions, or outcomes. If the
  memory does not contain the requested information, say so briefly.
- Be concise and specific, and answer the question directly. Do not add markdown or prose.

[USER]
Memory role: {{EPISODIC_OR_WORKING}}
{{PROBE_INSTRUCTION}}
\end{Verbatim}
\end{promptbox}
\captionof{figure}{Prompt used by the frozen policy to answer content-basis
probes from a fixed latent block.}
\label{fig:probe-generation}

\begin{center}
\small
\captionof{table}{Held-out content-basis probe templates. Aligned probes
target the same functional categories as training but use held-out wording
and response formats; Transfer probes query categories not used in training.}
\label{tab:probe-templates}
\begin{tabularx}{\textwidth}{@{}l l X X@{}}
\toprule
Category & Role & Information queried & Exact held-out instruction\\
\midrule
Semantic & Episodic & task, visual evidence, constraints, actions, outcome
& In 2--4 sentences, describe what task this trajectory pursues, the salient on-screen evidence, any constraints, the actions taken in order, and how it ended.\\
Semantic & Working & task, visible facts, actions, progress, latest state
& In 2--4 sentences, describe the current working state: the task, the important visible facts, the actions taken so far, the progress made, and the latest observed state.\\
Aligned & Episodic & reusable procedure and anchor decision
& State the reusable procedure this trajectory demonstrates and its single anchor decision --- the key action and the visible evidence that justified it.\\
Aligned & Working & latest state and observed change
& State the latest observed state and the state change that occurred across this working-memory segment.\\
Transfer & Episodic & remaining subgoal / harmful repetition
& Using only this memory, what subgoal still remains to finish the task, and is there an action already observed to fail that should not be repeated?\\
Transfer & Working & remaining subgoal / harmful repetition
& Using only this working memory, what subgoal remains before the task is done, and which recently attempted action failed and must not be repeated?\\
\bottomrule
\end{tabularx}
\end{center}

\subsubsection{Content-Basis Probe Judge}
\label{app:diag-judges}
The judge receives the raw chronological evidence, its memory role, the probe
instruction, and the candidate answer. It does not receive the method name,
training objective, checkpoint stage, or latent-token budget. Raw evidence is
the sole factual reference; the judge does not compare against a generated
teacher answer.
\begin{promptbox}{Content-basis probe judge}
\begin{Verbatim}[fontsize=\small,breaklines=true,breakanywhere=true]
[SYSTEM]
You are a STRICT, method-blind evaluator scoring how well a CANDIDATE ANSWER recovers
information from a GUI-agent trajectory. You are given the RAW chronological evidence
(screenshots and recorded actions) as the ONLY ground truth, the memory role, the probe
category, the probe instruction, and the candidate answer. You are NOT told which system,
training objective, checkpoint, or memory configuration produced the answer, and there is
NO reference answer -- judge the candidate only against the raw evidence.

Give one score from 0 to 100 (0 = wrong or entirely unsupported, 100 = fully correct,
complete, and grounded). Apply the criteria for the stated probe category:
- Semantic: factual correctness, coverage of the requested content, temporal consistency
  with the evidence, and absence of unsupported claims.
- Aligned: additionally reward correct recovery of the role-relevant target -- the reusable
  procedure and anchor decision for episodic memory, or the latest state and observed change
  for working memory.
- Transfer: whether the answer correctly identifies the remaining subgoal(s) and any harmful
  action repetition, without introducing unsupported advice.
Penalize any content not supported by the evidence. Be harsh: vague or partial answers score
in the middle, not high.

[USER]
Memory role: {{ROLE}}
Probe category: {{SEMANTIC_OR_ALIGNED_OR_TRANSFER}}
Probe instruction: {{PROBE_INSTRUCTION}}

Raw chronological evidence:
{{RAW_SCREENSHOTS_AND_ACTIONS}}

Candidate answer:
{{CANDIDATE_ANSWER}}

Return exactly:
{"score": <integer 0-100>, "reasoning": "<one sentence grounded in the raw evidence>"}
\end{Verbatim}
\end{promptbox}
\captionof{figure}{Method-blind judge used for Semantic, Aligned, and Transfer
content-basis probes.}
\label{fig:probe-judge}
Semantic probes are scored for factuality, coverage, temporal consistency,
and absence of unsupported claims. Aligned probes additionally emphasize
role-relevant procedure, anchor, state, or change recovery. Transfer probes
evaluate whether the answer correctly identifies remaining subgoals or
harmful repetition without introducing unsupported advice.

\subsubsection{One-Step Action-Validity Judge}
\label{app:action-judge}
The oracle-centered expansion and evidence-dependence intervention use the
same one-step judge. The judge evaluates whether the predicted action is a
valid next step rather than requiring exact agreement with a single reference
action.
\begin{promptbox}{One-step action-validity judge}
\begin{Verbatim}[fontsize=\small,breaklines=true,breakanywhere=true]
[SYSTEM]
You are a STRICT, method-blind evaluator of ONE GUI-agent action. You are given the task,
the recent action-observation history, the current screenshot, the available action schema,
and a single candidate next action. You are NOT told which system or memory setting produced
it, and you must NOT require agreement with any single reference action.

Decide whether the candidate action is a VALID next step. It is valid when it is (a) well-formed
under the action schema, (b) executable on the current screenshot -- it targets an element or
field that is actually visible, (c) consistent with the visible state and the task constraints,
and (d) a reasonable step toward completing the task. More than one action can be valid.

It is INVALID when it is malformed, targets an element that is not present, issues a stop answer
the current screen does not support, or repeats an action already observed to fail to change the
page. When genuinely uncertain, judge it invalid.

[USER]
Task instruction:
{{TASK}}

Recent action-observation history:
{{RECENT_HISTORY}}

Current screenshot:
{{CURRENT_SCREENSHOT}}

Available action schema:
{{ACTION_SCHEMA}}

Candidate next action:
{{PREDICTED_ACTION}}

Return exactly:
{"valid": true or false, "reasoning": "<one short sentence>"}
\end{Verbatim}
\end{promptbox}
\captionof{figure}{Method-blind one-step action-validity judge used in the
oracle-centered expansion and evidence-dependence intervention.}
\label{fig:action-judge}
A prediction is valid if it is executable in the current interface, consistent
with the visible state and task constraints, and constitutes a reasonable
next step toward completion. A valid action need not exactly match the
demonstrated action. Malformed actions, actions targeting absent elements,
unsupported stop answers, and repetitions of an action already observed to
fail are invalid. The judge does not receive the memory trajectory, intervention condition, or
Fixed/Dynamic identity.

\pdfcontentsdest{toc-d3}
\subsection{Diagnostic Construction and Configuration}
\label{app:diagconfig}

\subsubsection{Content-Basis Probe Sets}
\label{app:probe-sets}

We construct separate episodic and working-memory probe manifests from
a held-out WebVoyager pool. For each memory role, we randomly sample
tasks without replacement using a fixed random seed. At most one evidence item is retained
from any task, preventing tasks with longer trajectories or more
eligible decision points from receiving disproportionate weight.

The episodic manifest contains $200$ complete successful trajectories,
with one trajectory per sampled task. The working-memory manifest
contains $200$ expired history prefixes from a separately sampled set of
tasks. For a task with multiple eligible working-memory chunks, we uniformly
sample one non-empty chunk from its expired prefix and use that chunk as
the working-memory item. The episodic and working-memory
manifests are sampled independently and need not contain the same tasks.

Within each role, the same $200$ items are evaluated by all four
content-basis variants and all three probe categories. Thus, Semantic,
Aligned, and Transfer scores within a panel are paired at the item
level. The episodic and working-memory panels nevertheless use
role-specific probe instructions and are reported separately.

For every item, we construct its Stage-B fixed representation using the
corresponding role-specific base queries. Dynamic Readout and the trust
gate are disabled, and the frozen policy receives only the resulting
fixed latent block, the memory role, and the probe instruction. It does
not receive the raw trajectory, the current decision state, retrieved
context from other items, or a teacher-generated target. All variants
use identical item manifests, prompts, latent-token budgets, policy
parameters, generation settings, and judge inputs.

The six probe templates consist of one template for each
role--category pair. Semantic probes query general trajectory content.
Aligned probes query the functional information associated with the
corresponding role---reusable decisions, procedures, and outcomes for
episodic memory, and current state, observed changes, and progress for
working memory. Transfer probes query remaining subgoals and harmful
action repetition, which are not directly used as content-basis
training targets.

\subsubsection{Oracle-Centered Evidence Expansion}
\label{app:oracle-expansion}

\paragraph{Decision selection.}
We begin with WebVoyager episodes in which the Qwen3-VL-8B
No-Memory policy fails. From these episodes, we retain decision points
at which the generated next action is judged invalid and at which
external trajectory evidence can support at least one valid alternative.
We retain at most one decision point from each task. Decisions for which
no unambiguous supporting trajectory can be identified are discarded.
The resulting diagnostic contains $100$ decision points, and its
manifest is frozen before evaluating Fixed or Dynamic Readout.

\paragraph{Supporting-trajectory localization.}
For each retained decision, annotators inspect eligible trajectories in
the leakage-filtered episodic bank and identify one supporting
trajectory containing evidence applicable to the current decision.
They then mark the shortest contiguous fragment whose screenshots and
recorded actions provide sufficient evidence for at least one valid next
action. We call this fragment the \emph{Oracle Core}. The annotators do
not observe predictions from either Fixed or Dynamic Readout when
selecting the supporting trajectory or fragment boundaries. Cases with
multiple incompatible interpretations or no clearly sufficient fragment
are excluded.

This construction deliberately uses manually localized evidence and is
therefore not a retrieval benchmark. Its purpose is to hold the
decision-relevant evidence fixed while controlling how much surrounding
trajectory context competes for a bounded latent representation.

\paragraph{Evidence expansion.}
Let the selected supporting trajectory contain events
$\tau=(e_1,\ldots,e_T)$, and let its Oracle Core span the contiguous
interval $e_{\ell:r}$. We construct four nested evidence conditions:
\begin{align}
    S^{\mathrm{oracle}}
    &= e_{\ell:r}, \\
    S^{+2}
    &= e_{\max(1,\ell-1):\min(T,r+1)}, \\
    S^{+4}
    &= e_{\max(1,\ell-2):\min(T,r+2)}, \\
    S^{\mathrm{full}}
    &= e_{1:T}.
    \label{eq:oracle-expansion}
\end{align}
Thus, $+2$ adds at most one neighboring event on each side of the core,
whereas $+4$ adds at most two on each side. Expansion is clipped at the
trajectory boundaries and never padded with synthetic events.

\paragraph{Matched readout comparison.}
We train separate Fixed and Dynamic checkpoints for latent-token budgets
$B\in\{4,8,16\}$, where $B$ is the number of policy-facing tokens
produced for the single supporting memory block. For each budget, the
two readouts share the policy, evidence encoder, training data, action
supervision, optimization schedule, and latent-token budget; they differ
only in whether the compression queries depend on the current decision
state.

For every diagnostic decision, Fixed and Dynamic receive the same task,
recent history, current screenshot, working-memory input, supporting
trajectory, Oracle-Core boundaries, and evidence-expansion condition.
The trust gate is disabled, and all non-episodic policy inputs remain
unchanged across Oracle, $+2$, $+4$, and Full. Each model generates one
next action, which is evaluated by the method-blind one-step action-validity judge in
Fig.~\ref{fig:action-judge}. We report Action
Accuracy over the same $100$ decisions for every readout, evidence
condition, and latent budget.

\subsubsection{Evidence-Dependence Interventions}
\label{app:evidence-interventions}

We perform the evidence-dependence intervention on the same $100$
WebVoyager decisions used in the oracle-centered expansion. We fix one
Dynamic Readout checkpoint, disable the trust gate, and hold the task,
recent history, current screenshot, working memory, episodic slot count,
source embeddings, action prompt, and decoding configuration constant.
Only the information supplied through the three episodic memory slots is
changed.

For decision $d$, let $s_{d,j}$ denote the original trajectory assigned
to episodic slot $j\in\{1,2,3\}$, and let
\begin{equation}
    H_{d,j}=E_{\phi}(s_{d,j}),
    \qquad
    \bar h_{d,j}=\operatorname{Pool}(H_{d,j})
    \label{eq:evidence-intervention-representation}
\end{equation}
denote its token-level encoded evidence and pooled item summary,
respectively. These two representations serve different functions in
Dynamic Readout. The pooled summary conditions the residual-query
generator,
\begin{equation}
    Q_{d,j}
    =
    Q^{\mathrm{epi}}_0
    +
    A_{\eta}
    \bigl(
        u_d,\bar h_{d,j},e^{\mathrm{epi}}
    \bigr),
    \label{eq:evidence-intervention-query}
\end{equation}
whereas the resulting queries cross-attend to the full token-level
evidence:
\begin{equation}
    Z_{d,j}
    =
    C_{\Phi}
    \bigl(
        H_{d,j};Q_{d,j}
    \bigr)
    +
    e^{\mathrm{epi}}.
    \label{eq:evidence-intervention-readout}
\end{equation}
For this diagnostic, we refer to $\bar h_{d,j}$ as the
\emph{conditioning key} and to $H_{d,j}$ as the
\emph{evidence value}. These terms describe their functional roles in
the intervention and do not refer to the keys stored in the external
retrieval index.

Let $\pi$ be a fixed task-level derangement satisfying
$\pi(d)\neq d$. Donor evidence is drawn from other tasks in the same
benchmark domain and, where possible, from the same trajectory-length
bucket. Donors are disjoint from the current task under the leakage
checks in \S\ref{app:leakage}. The same permutation manifest is used
for every intervention condition. We compare the following four
settings.

\paragraph{Original Retrieval.}
Each episodic slot uses the matched conditioning key and evidence value
from its original retrieved trajectory:
\begin{equation}
    \mathcal I^{\mathrm{original}}_d
    =
    \left\{
        \bigl(
            \bar h_{d,j},
            H_{d,j}
        \bigr)
    \right\}_{j=1}^{3}.
    \label{eq:intervention-original}
\end{equation}
This is the unmodified Dynamic Readout condition.

\paragraph{No Evidence (State Only).}
We replace both the pooled item summary and the full encoded evidence
with zeros of the corresponding shapes:
\begin{equation}
    \mathcal I^{\mathrm{state}}_d
    =
    \left\{
        \bigl(
            \mathbf 0,
            \mathbf 0
        \bigr)
    \right\}_{j=1}^{3}.
    \label{eq:intervention-state-only}
\end{equation}
The residual-query generator and compressor are still executed, so the
episodic pathway retains the current decision state, learned base
queries, role identity, slot count, and slot ordering, but receives no
trajectory content. Working memory and the ordinary policy context
remain unchanged. ``State Only'' therefore refers specifically to the
episodic readout pathway rather than to the complete policy input.

\paragraph{Shuffled Retrieval.}
We replace each original item with a coherent conditioning-key--value
pair from the fixed same-domain donor task:
\begin{equation}
    \mathcal I^{\mathrm{shuffle}}_d
    =
    \left\{
        \bigl(
            \bar h_{\pi(d),j},
            H_{\pi(d),j}
        \bigr)
    \right\}_{j=1}^{3}.
    \label{eq:intervention-shuffled}
\end{equation}
The residual queries and the token-level evidence therefore describe
the same donor trajectory, but that trajectory is mismatched to the
current decision. This condition tests whether broadly applicable
procedural information remains useful even when retrieval selects the
wrong task.

\paragraph{Original Keys + Shuffled Values.}
We preserve the pooled summaries of the original retrieved items when
generating the residual queries, but replace the token-level evidence
read by the compressor with evidence from the fixed donor task:
\begin{equation}
    \mathcal I^{\mathrm{kv}}_d
    =
    \left\{
        \bigl(
            \bar h_{d,j},
            H_{\pi(d),j}
        \bigr)
    \right\}_{j=1}^{3}.
    \label{eq:intervention-key-value}
\end{equation}
Concretely, the query for slot $j$ is generated as
\begin{equation}
    Q^{\mathrm{kv}}_{d,j}
    =
    Q^{\mathrm{epi}}_0
    +
    A_{\eta}
    \bigl(
        u_d,\bar h_{d,j},e^{\mathrm{epi}}
    \bigr),
\end{equation}
but its policy-facing block is obtained by cross-attending to the
shuffled value:
\begin{equation}
    Z^{\mathrm{kv}}_{d,j}
    =
    C_{\Phi}
    \bigl(
        H_{\pi(d),j};Q^{\mathrm{kv}}_{d,j}
    \bigr)
    +
    e^{\mathrm{epi}}.
\end{equation}
This condition breaks the semantic correspondence between the item
summary used to determine what should be read and the trajectory
features from which the latent block is actually generated. Its
difference from Shuffled Retrieval is that Shuffled Retrieval retains a
coherent, although task-mismatched, summary--evidence pair, whereas this
condition deliberately mismatches the two components.

All four conditions preserve the number and ordering of episodic
blocks. They use the same checkpoint, decision states, working-memory
inputs, role embeddings, action prompt, and decoding configuration, and
the trust gate remains disabled throughout. We generate one action per
decision and report Action Accuracy under the common judge in
Fig.~\ref{fig:action-judge}. The judge receives no memory evidence,
intervention condition, or model identifier.

\subsubsection{Irrelevant-Trajectory Contamination}
\label{app:contamination}

\paragraph{Task subsets.}
We construct independent $100$-task subsets of WebVoyager and
Online-Mind2Web. These subsets are not used in the oracle-centered or
one-step evidence-dependence diagnostics. Each method performs a full
interactive rollout under the standard 15-step action budget and the
benchmark-specific success judge of \S\ref{app:judges}.

\paragraph{Nested contamination protocol.}
At each interaction step, the fixed retriever first produces the normal
top-three episodic candidates in descending similarity order,
\[
    \mathcal{R}_t
    =
    (s_{t,1},s_{t,2},s_{t,3}).
\]
We then replace exactly
$q\in\{0,1,2,3\}$ candidates with manually verified irrelevant
trajectories. Replacement is nested by retrieval rank:
\begin{align}
    q=0 &: \quad \text{no slot is replaced},\\
    q=1 &: \quad \text{replace rank }3,\\
    q=2 &: \quad \text{replace ranks }2\text{--}3,\\
    q=3 &: \quad \text{replace ranks }1\text{--}3.
    \label{eq:contamination-slots}
\end{align}
Thus, increasing $q$ retains all replacements made at lower
contamination levels and progressively removes the more highly ranked
original candidates. The corresponding contamination ratios are
$0\%$, $33\%$, $66\%$, and $100\%$.

For every task, step, and candidate slot, the irrelevant replacement is
selected before model evaluation and stored in a fixed contamination
manifest. Where possible, replacements are drawn from the same
benchmark domain so that they remain visually and procedurally
plausible, but they originate from unrelated tasks and are incompatible
with the current instruction and decision state. The same manifest is
used by all compared variants. Replaced trajectories are encoded and
processed normally; working memory and all unreplaced episodic
trajectories remain unchanged.

\paragraph{Compared variants.}
We evaluate three inference configurations:
\begin{itemize}
    \setlength{\itemsep}{0pt}
    \setlength{\parsep}{0pt}
    \setlength{\topsep}{2pt}

    \item \textbf{No Gate.}
    We use the full Stage-B checkpoint but bypass the trust decision at
    inference, forcing every non-empty episodic and working block to be
    admitted. The content basis and Dynamic Readout are otherwise
    identical to full FocusMem.

    \item \textbf{FocusMem.}
    We use the complete model and admit each block according to the
    validation-selected threshold $\gamma=0.3$.

    \item \textbf{Working-Only.}
    We remove all episodic blocks while preserving working memory and
    the ordinary decision context. Because no episodic evidence enters
    this configuration, its result is independent of the contamination
    level and serves as an episodic-free reference.
\end{itemize}

\paragraph{Success and gate passage.}
For WebVoyager and Online-Mind2Web, interactive SR is computed
separately with their benchmark-specific judges. Gate passage is
reported only for full FocusMem. At contamination level $q$, let
$m_{b,n,t,j}^{\mathrm{epi}}\in\{0,1\}$ indicate whether episodic slot
$j$ is admitted for benchmark $b$, task $n$, and executed step $t$.
We define
\begin{equation}
    \operatorname{Passage}(q)
    =
    100
    \times
    \frac{
        \displaystyle
        \sum_{b}
        \sum_{n}
        \sum_{t}
        \sum_{j=1}^{3}
        m_{b,n,t,j}^{\mathrm{epi}}
    }{
        \displaystyle
        \sum_{b}
        \sum_{n}
        \sum_{t}
        \sum_{j=1}^{3}
        \mathbb{I}
        \!\left[
            s_{b,n,t,j}^{\mathrm{epi}}
            \text{ is non-empty}
        \right]
    },
    \label{eq:gate-passage}
\end{equation}
where $b$ ranges over the WebVoyager and Online-Mind2Web
contamination subsets. The metric is therefore a pooled micro-average
over all non-empty episodic candidates, including both unreplaced and
injected trajectories. This definition also makes the clean
$q=0$ passage rate well-defined. Padding slots and missing candidates
are excluded from both numerator and denominator.

Table~\ref{tab:diagconfig} summarizes the configurations of the main
evaluation and selected diagnostic experiments. Unless noted, every
variant within an experiment shares the same policy, memory bank, retriever,
candidate set, and budgets, so differences are attributable to the compared
factor alone.

\begin{center}
\small
\captionof{table}{Evaluation configuration per experiment. All judgments use
Gemini-3.1-Pro (\S\ref{app:judges}). Table and figure references point to the
main paper.}
\label{tab:diagconfig}
% TODO(authors): confirm #Runs for the diagnostics (main results = 3).
\begin{tabularx}{\textwidth}{@{}>{\raggedright\arraybackslash}p{2.4cm} >{\raggedright\arraybackslash}p{3.0cm} >{\raggedright\arraybackslash}p{3.2cm} c >{\raggedright\arraybackslash}X@{}}
\toprule
Experiment & Data \& $N$ & Metric(s) & \#Runs & Key configuration\\
\midrule
Main results & 5 benchmarks, $N{=}308/$\allowbreak$200/$\allowbreak$150/$\allowbreak$598/$\allowbreak$266$ & SR, Overall & 1 & 15-step cap; all matched variants share bank, retriever, candidates, and budgets\\
{Ablation} & MMInA-Shopping, $N{=}200$ & SR, \#Steps, Hit-Max, Time/Task, Time/Step & 1 & cumulative: No Memory $\to$ +Fixed $\to$ +Content Basis $\to$ +Dynamic Readout (gate off) $\to$ +Gate\\
Contamination & WebVoyager and Online-Mind2Web, N=100+100& interactive SR; episodic-block passage rate & 1 & replace $0$--$3$ of the top-3 episodic candidates with high-confidence irrelevant negatives; variants no-gate / full / Working-Only\\
\bottomrule
\end{tabularx}
\end{center}

\pdfcontentsdest{toc-d4}
\subsection{Run Protocol}
\label{app:runs}

For the main results (Table~1 of the main paper), each
per-benchmark SR is computed from a single interactive
rollout of the full benchmark followed by judging.
The diagnostics in Table~3 and Figs.~3--4 are also
reported from a single run. We state \emph{one run} explicitly wherever a
number is not replicated rather than implying replication.

\pdfcontentsdest{toc-d5}
\subsection{Human Annotation Protocol}
\label{app:annotation}

For the contamination diagnostic (main paper, \S4.6), the injected episodic
trajectories are high-confidence irrelevant negatives---trajectories from
unrelated tasks, selected and checked by human annotators to be clearly
inapplicable to the current query, so that a robust trust gate should reject
them.

% ---- Software and hardware environment ----
\pdfcontentsdest{toc-e}
\section{Software and Hardware Environment}
\label{app:environment}

All experiments were conducted on a single server. Table~\ref{tab:software-environment}
lists the software stack used for training and evaluation, and
Table~\ref{tab:hardware-environment} summarizes the hardware configuration.

\begin{table}[H]
\centering
\small
\setlength{\tabcolsep}{5pt}
\begin{tabularx}{\linewidth}{@{}lX@{}}
\toprule
Software & Version\\
\midrule
Operating system & Ubuntu 22.04 LTS (Linux kernel 5.15.0-25-generic)\\
NVIDIA driver & 580.105.08\\
CUDA Driver API & 13.0\\
PyTorch & 2.12.1+cu130\\
Transformers & 4.57.3\\
vLLM & 0.21.0\\
Accelerate & 1.13.0\\
\bottomrule
\end{tabularx}
\caption{Software environment.}
\label{tab:software-environment}
\end{table}

\begin{table}[H]
\centering
\small
\setlength{\tabcolsep}{5pt}
\begin{tabularx}{\linewidth}{@{}lX@{}}
\toprule
Component & Configuration\\
\midrule
CPU & 2 \(\times\) Intel Xeon Gold 6530\\
CPU cores & 32 cores / 64 threads per socket; 64 physical cores / 128 logical
threads in total\\
CPU frequency & up to 4.0 GHz\\
CPU cache & L1d 3 MiB, L1i 2 MiB, L2 128 MiB, L3 320 MiB\\
NUMA & 4 NUMA nodes, each with \(\sim\)252 GiB memory and 32 logical CPUs\\
GPU & 8 \(\times\) NVIDIA RTX PRO 6000 Blackwell Server Edition\\
GPU memory & 96 GB per card (nominal); 97{,}887 MiB visible per card
(\(\sim\)95.6 GiB)\\
Total GPU memory & 768 GB (nominal); \(\sim\)764.7 GiB visible\\
GPU power cap & 600 W per card\\
GPU interface & PCIe 5.0 \(\times\)16 per card\\
GPU interconnect & GPU--GPU communication through PCIe/CPU NUMA\\
GPU P2P & P2P read/write supported on all GPU pairs; native atomic P2P not
supported\\
System memory & 1 TiB RAM; \(\sim\)1007.5 GiB detected\\
Swap & 68 GiB\\
\bottomrule
\end{tabularx}
\caption{Hardware environment.}
\label{tab:hardware-environment}
\end{table}

% ---- Training details and hyperparameter search ----
\pdfcontentsdest{toc-f}
\section{Training Details and Hyperparameter Search}
\label{app:training-details}

\pdfcontentsdest{toc-f1}
\subsection{Training Details}
\label{app:training-spec}

The Qwen3-VL-8B policy and the UGround-V1-7B grounding model are frozen
throughout training, and only the memory pathway is optimized. Action
supervision uses 23,438 successful high-quality trajectories of 8--15
interaction steps. Stage~B initializes from the Stage~A parameters and then
continues with a lower learning rate; Stages~A and~B use learning rates of
\(1\times10^{-5}\) and \(5\times10^{-6}\), respectively, with a global batch
size of 32. The memory objective combines the semantic and functional
content losses with \(\lambda_{\mathrm{sem}}=\lambda_{\mathrm{func}}=0.5\),
the content-basis weight \(\lambda_{\mathrm{basis}}=0.25\), and the trust-gate
weight \(\lambda_m=0.05\). The inference threshold \(\gamma=0.3\) is applied
to the trust score. Table~\ref{tab:training-hyperparameters} summarizes the
final training hyperparameters.

\begin{table}[H]
\centering
\small
\setlength{\tabcolsep}{5pt}
\begin{tabularx}{\linewidth}{@{}lX@{}}
\toprule
Hyperparameter & Setting\\
\midrule
Optimizer & AdamW\\
Learning-rate schedule & Cosine annealing with warmup ratio 0.1\\
Stage A learning rate & \(1\times10^{-5}\)\\
Stage B learning rate & \(5\times10^{-6}\)\\
Global batch size & 32\\
\(\lambda_{\mathrm{sem}}\) / \(\lambda_{\mathrm{func}}\) & 0.5 / 0.5\\
\(\lambda_{\mathrm{basis}}\) & 0.25\\
\(\lambda_m\) & 0.05\\
Trust threshold \(\gamma\) & 0.3\\
Gradient clipping & Max norm 1.0\\
Numerical precision & bfloat16\\
Distributed training & DDP\\
\bottomrule
\end{tabularx}
\caption{Final training hyperparameters.}
\label{tab:training-hyperparameters}
\end{table}

For evaluation, the frozen policy is served by vLLM with a maximum concurrency
of 32 and a context window of 32K tokens.

\pdfcontentsdest{toc-f2}
\subsection{Hyperparameter Search}
\label{app:hyperparameters}

In addition to the final settings above, we also considered alternative
choices during development: learning rates of \(1\times10^{-6}\),
\(5\times10^{-6}\), \(1\times10^{-5}\), and \(5\times10^{-5}\) for both stages;
global batch sizes of 16, 32, and 64 for both stages; and trust thresholds
\(\gamma\in\{0.1,0.2,0.3,0.4,0.5\}\). The settings in
Table~\ref{tab:training-hyperparameters} were selected on a held-out
validation split disjoint from the evaluation benchmarks.

% ============================================================
% Appendix section: Qualitative Case Studies (added by jjc)
% ============================================================
\pdfcontentsdest{toc-g}
\section{Qualitative Case Studies}
\label{app:cases}

We illustrate the three responsibilities of the factorized memory interface with
real evaluation trajectories. Each case is a genuine run of the same frozen
Qwen3-VL-8B policy under two memory conditions on one task; we show the actual
per-step screenshots and the model's action at each step, and never stage or edit
a trajectory. For a \texttt{click} action, the red dot marks the pixel that the
grounding model resolved the click \texttt{description} to, and the coordinate
$(x,y)$ is printed beneath that step in the screenshot's $1280\times720$ pixel
space. Coloured badges mark the step that demonstrates the mechanism. For
readability we show three to four representative steps per condition; long
identical stretches (repeated scrolls, searches, or clicks) are subsampled and
the omission is noted. All sites are local sandboxes, so screenshots contain no
personal data.

\pdfcontentsdest{toc-g1}
\subsection{Case 1: Content Recoverability (Episodic Reusable Experience)}
\label{app:case1}
\textbf{Task} (MMInA-Shopping): \emph{Which color is the body of the Coffee Mug Nordic Minimalist Coffee Cup Set European Small?} Gold answer: Blue.

\noindent\textbf{No Memory --- 15-step timeout, no answer (incorrect)}\par\vspace{2pt}

\begin{minipage}[t]{0.242\textwidth}\centering
{\setlength{\fboxsep}{0pt}\fbox{\includegraphics[width=\linewidth]{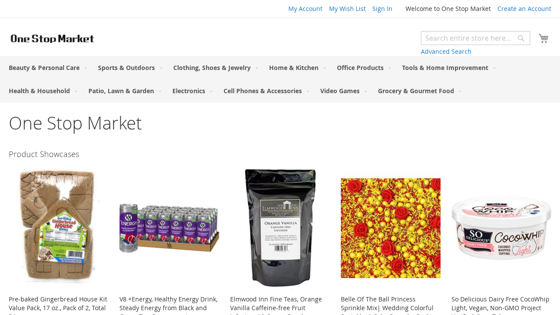}}}\\[2pt]
\parbox[t][6.5mm][t]{\linewidth}{\centering{\scriptsize\textbf{Step 1}}}\\[1pt]
\begin{Verbatim}[fontsize=\scriptsize,breaklines=true,breakanywhere=true]
{
  "name": "type",
  "arguments": {
    "text": "Coffee Mug Nordic Minimalist Coffee Cup Set European Small",
    "description": "the search bar at the top of the page"
  }
}
\end{Verbatim}
\end{minipage}\hfill
\begin{minipage}[t]{0.242\textwidth}\centering
{\setlength{\fboxsep}{0pt}\fbox{\includegraphics[width=\linewidth]{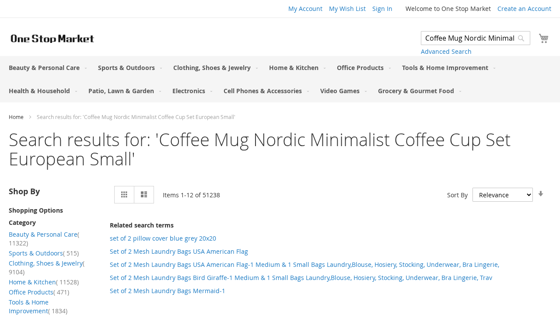}}}\\[2pt]
\parbox[t][6.5mm][t]{\linewidth}{\centering{\scriptsize\textbf{Step 2}}}\\[1pt]
\begin{Verbatim}[fontsize=\scriptsize,breaklines=true,breakanywhere=true]
{
  "name": "scroll",
  "arguments": {
    "direction": "down"
  }
}
\end{Verbatim}
\end{minipage}\hfill
\begin{minipage}[t]{0.242\textwidth}\centering
{\setlength{\fboxsep}{0pt}\fbox{\includegraphics[width=\linewidth]{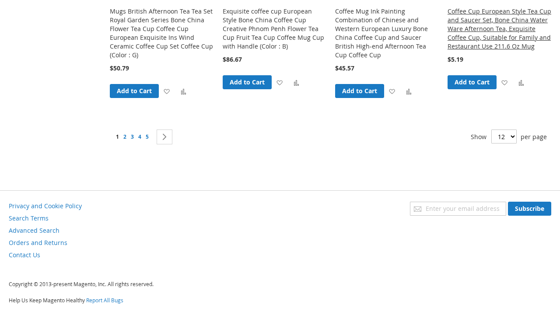}}}\\[2pt]
\parbox[t][6.5mm][t]{\linewidth}{\centering{\scriptsize\textbf{Step 14}}}\\[1pt]
\begin{Verbatim}[fontsize=\scriptsize,breaklines=true,breakanywhere=true]
{
  "name": "scroll",
  "arguments": {
    "direction": "down"
  }
}
\end{Verbatim}
\end{minipage}\hfill
\begin{minipage}[t]{0.242\textwidth}\centering
{\setlength{\fboxsep}{0pt}\fbox{\includegraphics[width=\linewidth]{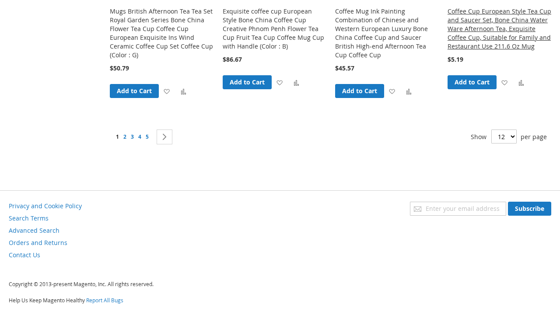}}}\\[2pt]
\parbox[t][6.5mm][t]{\linewidth}{\centering{\scriptsize\textbf{Step 15}}\\[1.5pt]\colorbox{red!70!black}{\textcolor{white}{\scriptsize\bfseries never opens the product}}}\\[1pt]
\begin{Verbatim}[fontsize=\scriptsize,breaklines=true,breakanywhere=true]
{
  "name": "scroll",
  "arguments": {
    "direction": "down"
  }
}
\end{Verbatim}
\end{minipage}
\par\vspace{5pt}

\noindent\textbf{FocusMem (latent) --- 5 steps, answer: Blue (correct)}\par\vspace{2pt}

\begin{minipage}[t]{0.242\textwidth}\centering
{\setlength{\fboxsep}{0pt}\fbox{\includegraphics[width=\linewidth]{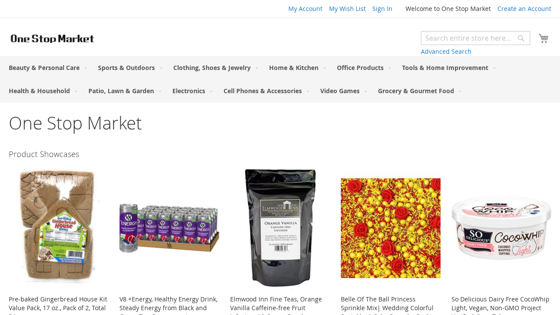}}}\\[2pt]
\parbox[t][6.5mm][t]{\linewidth}{\centering{\scriptsize\textbf{Step 1}}}\\[1pt]
\begin{Verbatim}[fontsize=\scriptsize,breaklines=true,breakanywhere=true]
{
  "name": "type",
  "arguments": {
    "text": "Coffee Mug Nordic Minimalist Coffee Cup Set European Small",
    "description": "the search box at the top right of the page with placeholder text 'Search entire store here...'"
  }
}
\end{Verbatim}
\end{minipage}\hfill
\begin{minipage}[t]{0.242\textwidth}\centering
{\setlength{\fboxsep}{0pt}\fbox{\includegraphics[width=\linewidth]{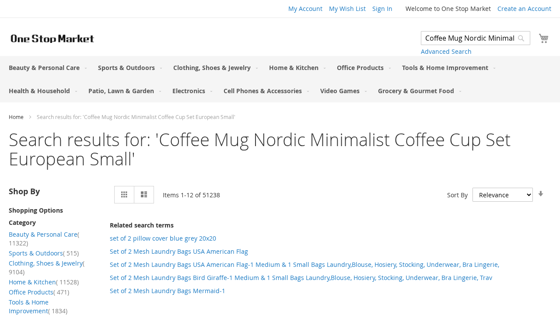}}}\\[2pt]
\parbox[t][6.5mm][t]{\linewidth}{\centering{\scriptsize\textbf{Step 2}}}\\[1pt]
\begin{Verbatim}[fontsize=\scriptsize,breaklines=true,breakanywhere=true]
{
  "name": "scroll",
  "arguments": {
    "direction": "down"
  }
}
\end{Verbatim}
\end{minipage}\hfill
\begin{minipage}[t]{0.242\textwidth}\centering
{\setlength{\fboxsep}{0pt}\fbox{\includegraphics[width=\linewidth]{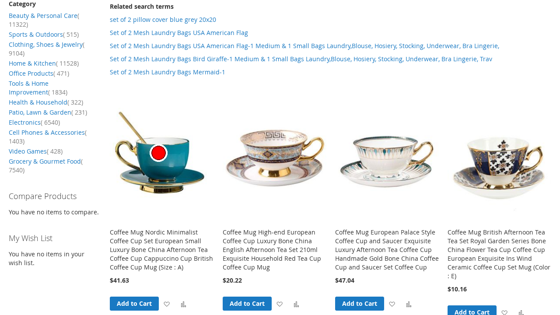}}}\\[2pt]
\parbox[t][6.5mm][t]{\linewidth}{\centering{\scriptsize\textbf{Step 3}}\\[1.5pt]\colorbox{green!55!black}{\textcolor{white}{\scriptsize\bfseries recovers: opens the product}}}\\[1pt]
\begin{Verbatim}[fontsize=\scriptsize,breaklines=true,breakanywhere=true]
{
  "name": "click",
  "arguments": {
    "description": "the first product image showing a blue coffee mug with gold trim and a saucer"
  }
}
\end{Verbatim}
{\scriptsize\itshape red dot: grounding $(x,y)=(362,350)$}
\end{minipage}\hfill
\begin{minipage}[t]{0.242\textwidth}\centering
{\setlength{\fboxsep}{0pt}\fbox{\includegraphics[width=\linewidth]{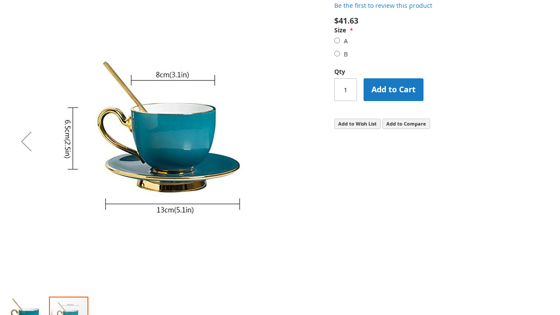}}}\\[2pt]
\parbox[t][6.5mm][t]{\linewidth}{\centering{\scriptsize\textbf{Step 5}}}\\[1pt]
\begin{Verbatim}[fontsize=\scriptsize,breaklines=true,breakanywhere=true]
{
  "name": "stop",
  "arguments": {
    "answer": "Blue"
  }
}
\end{Verbatim}
\end{minipage}
\par\vspace{5pt}

\captionof{figure}{Case~1. Same frozen Qwen3-VL-8B policy under No Memory (top) and FocusMem (bottom). Steps 3--14 of the No-Memory run repeat scrolls of the listing and are subsampled.}
\label{fig:case1}

\noindent Both agents share the same frozen policy and 15-step budget. Under No Memory the agent searches the product, reaches the results listing, and then scrolls that listing for the entire budget without ever opening the product page, timing out with no answer. FocusMem opens the product on its third step (red dot on the product image): the retrieved episodic memory at that step is a sibling coffee-cup task whose logged action is exactly a click on the first product image, supplying the reusable ``open the product to read its attribute'' procedure that No Memory never performs. It then reads the true teal-blue body and answers in five steps. The contrast isolates content recoverability for episodic memory---the compressed latent block retained a reusable procedure, not merely a surface action.

\pdfcontentsdest{toc-g2}
\subsection{Case 2: Working Memory Retains Progress}
\label{app:case2}
\textbf{Task} (MMInA-Shopping): \emph{Which one has a higher heel, the Amarantos Women's D'Orsay Pumps Stiletto High Heel Shoes or the Birkenstock unisex-adult Contains a Toe?} Gold answer: Amarantos.

\noindent\textbf{No Memory --- 15-step timeout, no answer (incorrect)}\par\vspace{2pt}

\begin{minipage}[t]{0.242\textwidth}\centering
{\setlength{\fboxsep}{0pt}\fbox{\includegraphics[width=\linewidth]{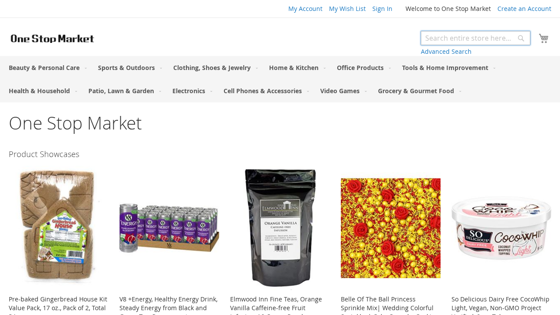}}}\\[2pt]
\parbox[t][6.5mm][t]{\linewidth}{\centering{\scriptsize\textbf{Step 2}}}\\[1pt]
\begin{Verbatim}[fontsize=\scriptsize,breaklines=true,breakanywhere=true]
{
  "name": "type",
  "arguments": {
    "text": "Amarantos Women's D'Orsay Pumps Stiletto High Heel Shoes",
    "description": "the search bar at the top of the page"
  }
}
\end{Verbatim}
\end{minipage}\hfill
\begin{minipage}[t]{0.242\textwidth}\centering
{\setlength{\fboxsep}{0pt}\fbox{\includegraphics[width=\linewidth]{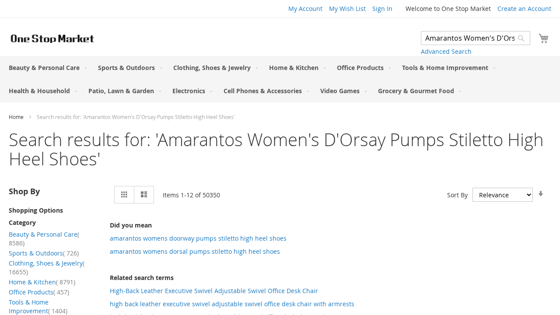}}}\\[2pt]
\parbox[t][6.5mm][t]{\linewidth}{\centering{\scriptsize\textbf{Step 3}}}\\[1pt]
\begin{Verbatim}[fontsize=\scriptsize,breaklines=true,breakanywhere=true]
{
  "name": "scroll",
  "arguments": {
    "direction": "down"
  }
}
\end{Verbatim}
\end{minipage}\hfill
\begin{minipage}[t]{0.242\textwidth}\centering
{\setlength{\fboxsep}{0pt}\fbox{\includegraphics[width=\linewidth]{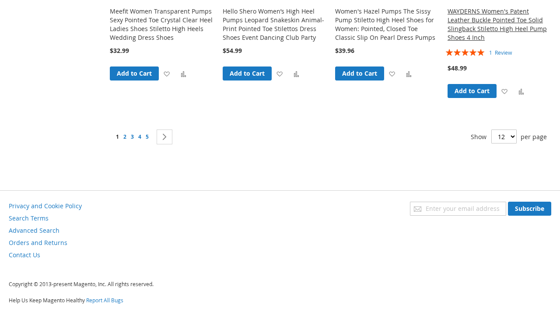}}}\\[2pt]
\parbox[t][6.5mm][t]{\linewidth}{\centering{\scriptsize\textbf{Step 14}}}\\[1pt]
\begin{Verbatim}[fontsize=\scriptsize,breaklines=true,breakanywhere=true]
{
  "name": "scroll",
  "arguments": {
    "direction": "down"
  }
}
\end{Verbatim}
\end{minipage}\hfill
\begin{minipage}[t]{0.242\textwidth}\centering
{\setlength{\fboxsep}{0pt}\fbox{\includegraphics[width=\linewidth]{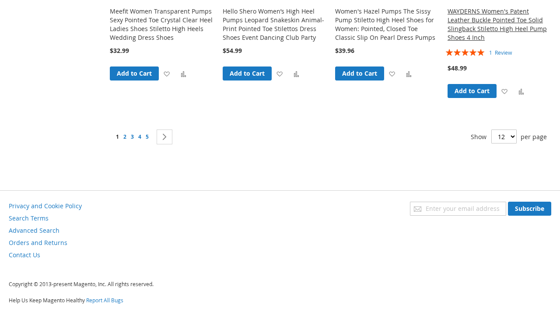}}}\\[2pt]
\parbox[t][6.5mm][t]{\linewidth}{\centering{\scriptsize\textbf{Step 15}}\\[1.5pt]\colorbox{red!70!black}{\textcolor{white}{\scriptsize\bfseries never reaches the products}}}\\[1pt]
\begin{Verbatim}[fontsize=\scriptsize,breaklines=true,breakanywhere=true]
{
  "name": "scroll",
  "arguments": {
    "direction": "down"
  }
}
\end{Verbatim}
\end{minipage}
\par\vspace{5pt}

\noindent\textbf{FocusMem (latent) --- 15 steps, answer: Amarantos (correct)}\par\vspace{2pt}

\begin{minipage}[t]{0.242\textwidth}\centering
{\setlength{\fboxsep}{0pt}\fbox{\includegraphics[width=\linewidth]{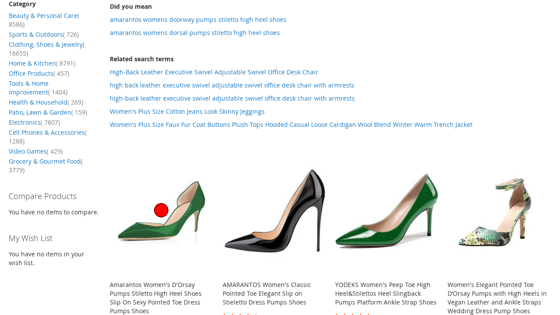}}}\\[2pt]
\parbox[t][6.5mm][t]{\linewidth}{\centering{\scriptsize\textbf{Step 3}}\\[1.5pt]\colorbox{green!55!black}{\textcolor{white}{\scriptsize\bfseries sees product A (high heel)}}}\\[1pt]
\begin{Verbatim}[fontsize=\scriptsize,breaklines=true,breakanywhere=true]
{
  "name": "click",
  "arguments": {
    "description": "the first product image showing green Amarantos Women's D'Orsay Pumps Stiletto High Heel Shoes"
  }
}
\end{Verbatim}
{\scriptsize\itshape red dot: grounding $(x,y)=(369,481)$}
\end{minipage}\hfill
\begin{minipage}[t]{0.242\textwidth}\centering
{\setlength{\fboxsep}{0pt}\fbox{\includegraphics[width=\linewidth]{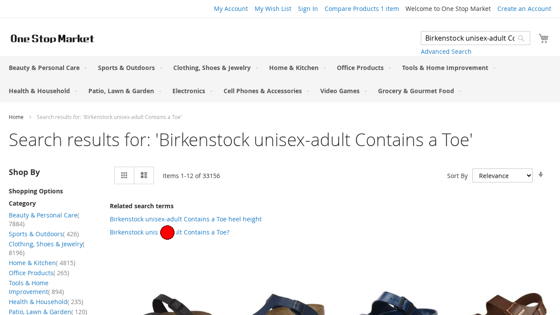}}}\\[2pt]
\parbox[t][6.5mm][t]{\linewidth}{\centering{\scriptsize\textbf{Step 8}}}\\[1pt]
\begin{Verbatim}[fontsize=\scriptsize,breaklines=true,breakanywhere=true]
{
  "name": "click",
  "arguments": {
    "description": "the first product image showing Birkenstock unisex-adult Contains a Toe shoes in the search results"
  }
}
\end{Verbatim}
{\scriptsize\itshape red dot: grounding $(x,y)=(383,531)$}
\end{minipage}\hfill
\begin{minipage}[t]{0.242\textwidth}\centering
{\setlength{\fboxsep}{0pt}\fbox{\includegraphics[width=\linewidth]{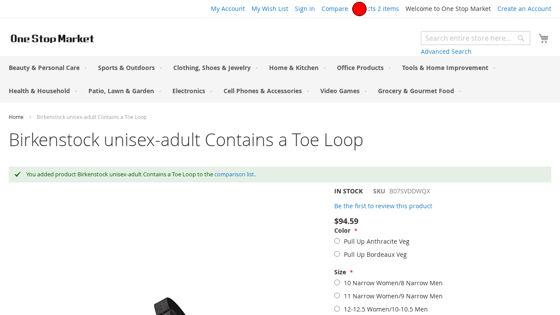}}}\\[2pt]
\parbox[t][6.5mm][t]{\linewidth}{\centering{\scriptsize\textbf{Step 13}}\\[1.5pt]\colorbox{green!55!black}{\textcolor{white}{\scriptsize\bfseries recalls A to compare}}}\\[1pt]
\begin{Verbatim}[fontsize=\scriptsize,breaklines=true,breakanywhere=true]
{
  "name": "click",
  "arguments": {
    "description": "the 'Compare Products 2 items' link in the top navigation bar"
  }
}
\end{Verbatim}
{\scriptsize\itshape red dot: grounding $(x,y)=(822,20)$}
\end{minipage}\hfill
\begin{minipage}[t]{0.242\textwidth}\centering
{\setlength{\fboxsep}{0pt}\fbox{\includegraphics[width=\linewidth]{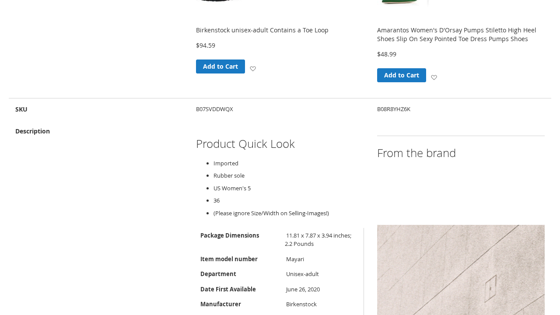}}}\\[2pt]
\parbox[t][6.5mm][t]{\linewidth}{\centering{\scriptsize\textbf{Step 15}}}\\[1pt]
\begin{Verbatim}[fontsize=\scriptsize,breaklines=true,breakanywhere=true]
{
  "name": "stop",
  "arguments": {
    "answer": "Amarantos Women's D'Orsay Pumps Stiletto High Heel Shoes"
  }
}
\end{Verbatim}
\end{minipage}
\par\vspace{5pt}

\captionof{figure}{Case~2. A two-product comparison under No Memory (top) and FocusMem (bottom). Green badges trace how FocusMem stores product~A and recalls it; repeated scrolls and intermediate navigation are subsampled to four steps.}
\label{fig:case2}

\noindent A comparison task must retain the first product's attribute while it navigates away to inspect the second. No Memory searches the first product but never reaches either product page---it scrolls the listing to the step cap and times out, so it never even acquires product~A's heel to compare. FocusMem instead opens product~A (the Amarantos stiletto), adds it to the site's compare list, then searches and opens product~B (the flat Birkenstock sandal), opens the two-item comparison view, and answers ``Amarantos.'' The final decision depends on the product-A observation made ten steps earlier, beyond the recent-history window; the compressed working memory of the expired prefix carries that earlier progress forward.

\pdfcontentsdest{toc-g3}
\subsection{Case 3: State-Conditioned Readout (Fixed vs.\ Dynamic Query)}
\label{app:case3}
\textbf{Task} (MMInA-Wikipedia): \emph{Are there buildings in Nanning that look like a golden sphere?} Gold answer: No. Controlled A/B on two checkpoints differing only in the query mechanism, with greedy decoding, so any divergence is attributable to readout. The divergent actions are a re-typed search versus a scroll---no grounding is involved.

\noindent\textbf{Fixed-query readout --- 15-step re-search loop, no answer (incorrect)}\par\vspace{2pt}

\begin{minipage}[t]{0.242\textwidth}\centering
{\setlength{\fboxsep}{0pt}\fbox{\includegraphics[width=\linewidth]{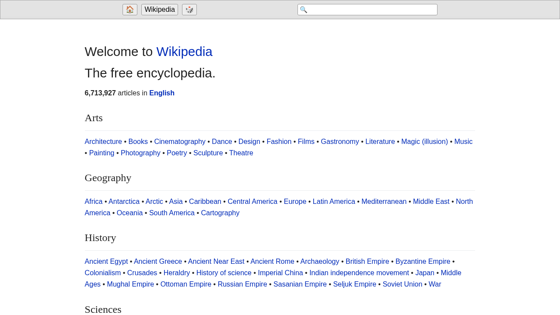}}}\\[2pt]
\parbox[t][6.5mm][t]{\linewidth}{\centering{\scriptsize\textbf{Step 1}}}\\[1pt]
\begin{Verbatim}[fontsize=\scriptsize,breaklines=true,breakanywhere=true]
{
  "name": "type",
  "arguments": {
    "text": "Nanning golden sphere building",
    "description": "the search box at the top of the Wikipedia page"
  }
}
\end{Verbatim}
\end{minipage}\hfill
\begin{minipage}[t]{0.242\textwidth}\centering
{\setlength{\fboxsep}{0pt}\fbox{\includegraphics[width=\linewidth]{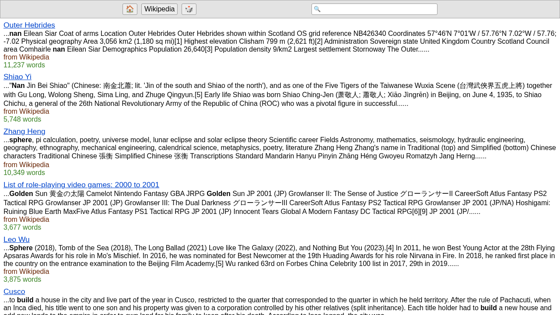}}}\\[2pt]
\parbox[t][6.5mm][t]{\linewidth}{\centering{\scriptsize\textbf{Step 4}}\\[1.5pt]\colorbox{red!70!black}{\textcolor{white}{\scriptsize\bfseries re-types the same search}}}\\[1pt]
\begin{Verbatim}[fontsize=\scriptsize,breaklines=true,breakanywhere=true]
{
  "name": "type",
  "arguments": {
    "text": "Nanning golden sphere building",
    "description": "the search box at the top of the Wikipedia page"
  }
}
\end{Verbatim}
\end{minipage}\hfill
\begin{minipage}[t]{0.242\textwidth}\centering
{\setlength{\fboxsep}{0pt}\fbox{\includegraphics[width=\linewidth]{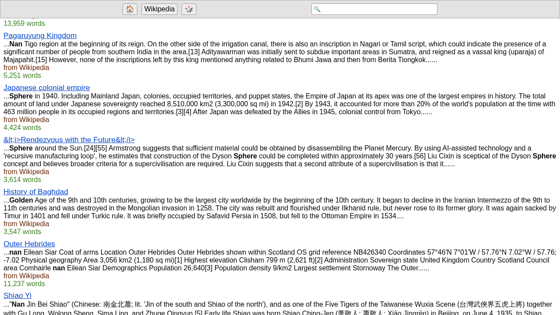}}}\\[2pt]
\parbox[t][6.5mm][t]{\linewidth}{\centering{\scriptsize\textbf{Step 10}}}\\[1pt]
\begin{Verbatim}[fontsize=\scriptsize,breaklines=true,breakanywhere=true]
{
  "name": "type",
  "arguments": {
    "text": "Nanning golden sphere building",
    "description": "the search box at the top of the Wikipedia page"
  }
}
\end{Verbatim}
\end{minipage}\hfill
\begin{minipage}[t]{0.242\textwidth}\centering
{\setlength{\fboxsep}{0pt}\fbox{\includegraphics[width=\linewidth]{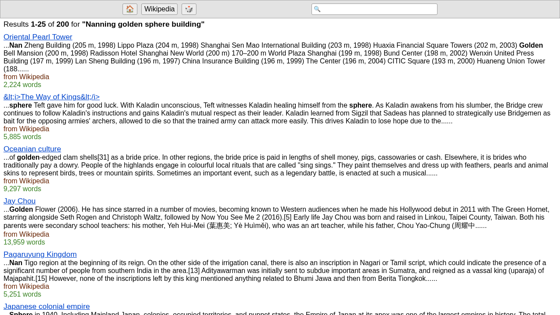}}}\\[2pt]
\parbox[t][6.5mm][t]{\linewidth}{\centering{\scriptsize\textbf{Step 15}}\\[1.5pt]\colorbox{red!70!black}{\textcolor{white}{\scriptsize\bfseries still re-searching; no answer}}}\\[1pt]
\begin{Verbatim}[fontsize=\scriptsize,breaklines=true,breakanywhere=true]
{
  "name": "type",
  "arguments": {
    "text": "Nanning golden sphere building",
    "description": "the search box at the top of the Wikipedia page"
  }
}
\end{Verbatim}
\end{minipage}
\par\vspace{5pt}

\noindent\textbf{Dynamic-query readout --- 15 steps, answer: No (correct)}\par\vspace{2pt}

\begin{minipage}[t]{0.242\textwidth}\centering
{\setlength{\fboxsep}{0pt}\fbox{\includegraphics[width=\linewidth]{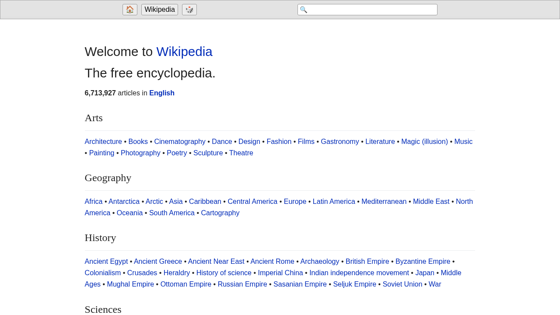}}}\\[2pt]
\parbox[t][6.5mm][t]{\linewidth}{\centering{\scriptsize\textbf{Step 1}}}\\[1pt]
\begin{Verbatim}[fontsize=\scriptsize,breaklines=true,breakanywhere=true]
{
  "name": "type",
  "arguments": {
    "text": "Nanning golden sphere building",
    "description": "the search box at the top of the Wikipedia page"
  }
}
\end{Verbatim}
\end{minipage}\hfill
\begin{minipage}[t]{0.242\textwidth}\centering
{\setlength{\fboxsep}{0pt}\fbox{\includegraphics[width=\linewidth]{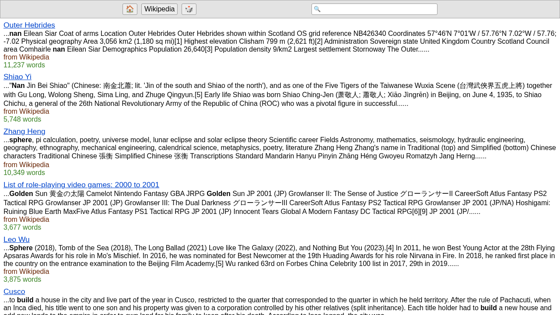}}}\\[2pt]
\parbox[t][6.5mm][t]{\linewidth}{\centering{\scriptsize\textbf{Step 4}}\\[1.5pt]\colorbox{green!55!black}{\textcolor{white}{\scriptsize\bfseries scrolls to read the page}}}\\[1pt]
\begin{Verbatim}[fontsize=\scriptsize,breaklines=true,breakanywhere=true]
{
  "name": "scroll",
  "arguments": {
    "direction": "down"
  }
}
\end{Verbatim}
\end{minipage}\hfill
\begin{minipage}[t]{0.242\textwidth}\centering
{\setlength{\fboxsep}{0pt}\fbox{\includegraphics[width=\linewidth]{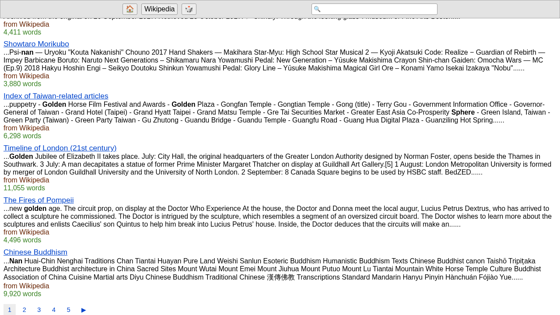}}}\\[2pt]
\parbox[t][6.5mm][t]{\linewidth}{\centering{\scriptsize\textbf{Step 14}}}\\[1pt]
\begin{Verbatim}[fontsize=\scriptsize,breaklines=true,breakanywhere=true]
{
  "name": "scroll",
  "arguments": {
    "direction": "down"
  }
}
\end{Verbatim}
\end{minipage}\hfill
\begin{minipage}[t]{0.242\textwidth}\centering
{\setlength{\fboxsep}{0pt}\fbox{\includegraphics[width=\linewidth]{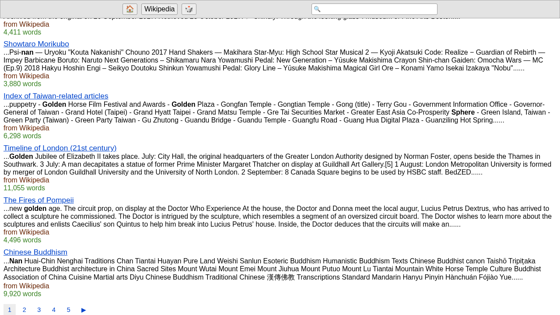}}}\\[2pt]
\parbox[t][6.5mm][t]{\linewidth}{\centering{\scriptsize\textbf{Step 15}}\\[1.5pt]\colorbox{green!55!black}{\textcolor{white}{\scriptsize\bfseries reads it, answers No}}}\\[1pt]
\begin{Verbatim}[fontsize=\scriptsize,breaklines=true,breakanywhere=true]
{
  "name": "stop",
  "arguments": {
    "answer": "No"
  }
}
\end{Verbatim}
\end{minipage}
\par\vspace{5pt}

\captionof{figure}{Case~3. Matched checkpoints trained under identical settings
except for the readout design. At step~4 both stand on the same Nanning search-results page with the same retrieval set, but the fixed-query readout re-issues the identical search query (and keeps re-typing it), whereas the dynamic-query readout scrolls to read the page. The intermediate re-typing / scrolling steps are subsampled.}
\label{fig:case3}

\noindent At step~4 both variants stand on the identical Nanning search-results page with the same top-3 retrieval set. From that shared state the fixed-query readout re-issues the very same search query and keeps re-typing it for the rest of the budget---it never settles to read a page and times out with no answer. The state-conditioned dynamic readout instead scrolls through the page to read its content and answers ``No'' correctly. The divergent actions are a re-typed \texttt{type} versus a \texttt{scroll}, so no grounding is involved and the difference reflects the readout variant rather than
grounding or decoding. Consistent with the claim that state-conditioned readout is more robust rather than always superior, across the eleven clean fixed-vs-dynamic divergences the dynamic readout converts three failures to successes and the fixed two; this case is one of the dynamic wins.

\end{document}